\documentclass{article}

\PassOptionsToPackage{numbers, compress}{natbib}
    \usepackage[final]{neurips_2025}

\usepackage[utf8]{inputenc} % allow utf-8 input
\usepackage[T1]{fontenc}    % use 8-bit T1 fonts
\usepackage{hyperref}       % hyperlinks
\usepackage{url}            % simple URL typesetting
\usepackage{booktabs}       % professional-quality tables
\usepackage{amsfonts}       % blackboard math symbols
\usepackage{nicefrac}       % compact symbols for 1/2, etc.
\usepackage{microtype}      % microtypography
\usepackage{amsmath}

\usepackage{hyperref}
\usepackage[table,xcdraw]{xcolor}
\usepackage{colortbl}
\usepackage{multirow}
\usepackage{makecell}
\usepackage{booktabs}
\usepackage{array}
\usepackage{graphicx}
\usepackage{subcaption}
\usepackage{wrapfig}
\usepackage{listings}

\usepackage{pifont}

\RequirePackage{algorithm}
\RequirePackage{algorithmic}

\newcommand{\thinmidrule}{\arrayrulecolor{black!30}\specialrule{0.05pt}{0pt}{0pt}}
\newcommand{\regmidrule}{\arrayrulecolor{black!100}\midrule}

\newcommand\dex{\textsc{Dex}}
\newcommand\abbr{\textsc{Diff}}
\newcommand\diff{\textsc{Diff} Transformer}

\definecolor{win}{RGB}{185, 246, 202}
\definecolor{lose}{RGB}{238, 238, 238}

\definecolor{deltaGreen}{RGB}{0, 128, 0}
\definecolor{deltaRed}{RGB}{220, 20, 60}
\definecolor{royalblue}{RGB}{65,105,225}
\definecolor{mediumseagreen}{RGB}{60,179,113}
\definecolor{tomato}{RGB}{255,99,71}
\definecolor{plotblue}{RGB}{0,0,255}
\definecolor{plotred}{RGB}{255,0,0}
\definecolor{viridislight}{RGB}{136,209,109}

\definecolor{codegreen}{rgb}{0,0.6,0}
\definecolor{codegray}{rgb}{0.5,0.5,0.5}
\definecolor{codepurple}{rgb}{0.58,0,0.82}
\definecolor{backcolour}{rgb}{1,1,1}

\definecolor{jiho}{rgb}{0, 0, 0.8}

\lstdefinestyle{mystyle}{
    backgroundcolor=\color{backcolour},   
    commentstyle=\color{codegreen},
    keywordstyle=\color{magenta},
    numberstyle=\tiny\color{codegray},
    stringstyle=\color{codepurple},
    basicstyle=\ttfamily\footnotesize,
    breakatwhitespace=true,         
    breaklines=true,                 
    captionpos=b,                    
    keepspaces=true,                 
    numbersep=5pt,                  
    showspaces=false,                
    showstringspaces=false,
    showtabs=false,                  
    tabsize=2,
}

\title{Understanding Differential Transformer \protect\\Unchains Pretrained Self-Attentions}

\author{
\begin{minipage}{\textwidth}
    \centering
    Chaerin Kong$^{1,2}{}^*$ \hspace{1.25cm} Jiho Jang$^2{}\thanks{Equal contributions.}$ \hspace{1.25cm} Nojun Kwak$^{2}$
    \vspace{2mm}
\end{minipage}\\
\textsuperscript{1} TwelveLabs
\hspace{4mm}
\textsuperscript{2} Seoul National University
\vspace{2mm}
\\
\hspace{-8mm}
    {\tt \href{mailto:chaerin.k.kong@gmail.com}{chaerin.k.kong@gmail.com}}
}
\begin{document}

\maketitle

\vspace{-1.5em}
\begin{abstract}
  Differential Transformer has recently gained significant attention for its impressive empirical performance, often attributed to its ability to perform noise canceled attention. However, precisely how differential attention achieves its empirical benefits remains poorly understood. Moreover, Differential Transformer architecture demands large-scale training from scratch, hindering utilization of open pretrained weights. In this work, we conduct an in-depth investigation of Differential Transformer, uncovering three key factors behind its success: (1) enhanced expressivity via negative attention, (2) reduced redundancy among attention heads, and (3) improved learning dynamics. 
  Based on these findings, we propose \dex, a novel method to efficiently integrate the advantages of differential attention into pretrained language models.
  By reusing the softmax attention scores and adding a lightweight differential operation on the output value matrix, \dex~effectively incorporates the key advantages of differential attention while remaining lightweight in both training and inference.
  Evaluations confirm that \dex~substantially improves the pretrained LLMs across diverse benchmarks, achieving significant performance gains with minimal adaptation data (< 0.01\%).
  % Thorough evaluation demonstrates that our approach successfully bridges the gap between traditional and Differential Transformer, achieving substantial performance gain with less than 0.01\% of the full training data.

% talk more about our extension method

\end{abstract}

\section{Introduction}

Transformer-based architectures have emerged as the cornerstone of modern deep learning across multiple domains~\cite{vaswani2017attention,dosovitskiy2020image,peebles2023scalable,chen2021decision,radford2021learning,brown2020language,jang2023unifying,kirillov2023segment,carion2020end,devlin2019bert,kong2023leveraging,jang2023self}. With their attention mechanism, transformers effectively model long-range dependencies, leading to significant advances in large language models~\cite{touvron2023llama,bai2023qwen,liu2024deepseek,team2025gemma,abdin2024phi,brown2020language}. However, a growing body of work~\cite{LLMTest_NeedleInAHaystack,liu2024lost,lu2021fantastically} highlights that these language models struggle with key information retrieval due to inherent \textit{attention noise}.

To address this issue, Differential (\abbr) Transformer~\cite{ye2024differential} introduces differential attention that computes the difference between two attention scores, thereby boosting attention on key tokens while suppressing common noise. Although its strong empirical performance has established it as a promising alternative to standard transformers, 
how this simple architecture consistently harnesses the differential operation for effective noise cancellation without explicit guidance remains elusive.
Moreover, due to the gap in architecture, employing \abbr~attention requires training from scratch, which prohibits utilization of open pretrained weights~\cite{touvron2023llama,bai2023qwen,team2025gemma,liu2024deepseek,mistral7b_announcement,groeneveld2024olmo,abdin2024phi} and incurs huge cost.

In this paper, we aim to fill this gap by providing an in-depth analysis of the mechanisms of \diff~and presenting a method to efficiently integrate its benefits into existing pretrained transformers. Our key observations are threefold. (1) \abbr~attention enhances expressivity through negative attention scores. (2) \abbr~attention reduces redundancy among its attention heads. (3) \diff~exhibits improved learning dynamics.

Building on these insights, we present \dex~(\textbf{D}ifferential \textbf{Ex}tension), an efficient framework that injects the strengths of \diff~into a pretrained LLM without training from scratch.
Unlike most finetuning methods that fit the model to downstream data, \dex~is an architectural adaptation strategy that introduces a
key mechanism from a different architecture to a pretrained model,
conceptually similar to MHA2MLA~\cite{ji2025towards}.
% Specifically, \dex~introduces a learnable differential mechanism to the the output value matrix (softmax$(\mathbf{QK}^T)\mathbf{V}$) instead of the attention score (softmax$(\mathbf{QK}^T)$), making the adaptation lightweight (both training and inference) yet nonetheless effective, as demonstrated empirically.
Specifically, \dex~operates by reusing the pretrained softmax attention scores (softmax$(\mathbf{QK}^T)$) and applying its learnable differential mechanism to the output value matrix (softmax$(\mathbf{QK}^T)\mathbf{V}$), making the adaptation lightweight (in both training and inference) yet effective, as demonstrated empirically.
% Specifically, \dex~applies differential mechanism to the output value matrix (softmax$(\mathbf{QK}^T)\mathbf{V}$) instead of the attention score (softmax$(\mathbf{QK}^T)$), which lets us fully leverage the pretrained attention without the need for retraining or additional inference compute.
To facilitate stable and performant transition, we introduce additional techniques for head selection and $\lambda$-annealing, which controls the critical balance between original knowledge and incoming architectural changes.
We validate \dex~across multiple model families (Llama-3~\cite{grattafiori2024llama} and Qwen-2.5~\cite{yang2024qwen2}) and scales (0.5B-8B), using diverse benchmarks such as language modeling~\cite{eval-harness,wang2018glue,wang2019superglue}, key information retrieval~\cite{LLMTest_NeedleInAHaystack} and in-context learning~\cite{bertsch2024context}. 
\dex~consistently achieves significant gains using less than 0.01\% the size of the original training data (<1B tokens), without incurring nontrivial test-time overhead.
% Our empirical results demonstrate the consistent effectiveness and versatility of the proposed method.
% Our approach efficiently integrates differential mechanism into pretrained attentions by reusing the original attention, only adding a learnable projection for the value matrix instead of retraining or recomputing the attention itself.

\section{How Does Differential Transformer Work?}
\label{sec:2}

In this section, we systematically analyze the internal mechanics of \diff. Since the original weights are not publicly available at the time of writing, we train a \diff~on a similar data mix to carry out our analyses. Please refer to Appendix \ref{supp:5.2} for full details.

\subsection{Preliminary: \diff}
\label{subsec: prelim}

The key innovation of \diff~is replacing the softmax attentions with \abbr~attentions. \abbr~attention introduces a mechanism designed to suppress attention noise by computing the difference between attention scores from two separate \textit{groups}. Given an input sequence $X \in \mathbb{R}^{N \times d_{\text{model}}}$, it is first projected into queries, keys, and values as follows:
\begin{equation}
    [Q_1; Q_2] = XW_Q, \quad [K_1; K_2] = XW_K, \quad V = XW_V,
\end{equation}
where $Q_1, Q_2, K_1, K_2 \in \mathbb{R}^{N \times d}$ and $V \in \mathbb{R}^{N \times 2d}$ denote projected matrices, and $W_Q, W_K, W_V \in \mathbb{R}^{d_{\text{model}} \times 2d}$ are learnable parameters. The differential attention is then computed as:
\begin{equation}
    \text{DiffAttn}(X) = \left( 
    \text{softmax}\left(\frac{Q_1 K_1^\top}{\sqrt{d}}\right) 
    - \lambda \cdot \text{softmax}\left(\frac{Q_2 K_2^\top}{\sqrt{d}}\right) 
    \right) V,
\end{equation}
where $\lambda$ is a learnable scalar. This differential mechanism enhances robustness by canceling common-mode attention noise, similar in spirit to differential amplifiers. 
We note that despite \abbr~attention having the same number of parameters, it exhibits significantly higher compute cost and peak memory usage in practice due to enlarged dimensions (see Fig.\ref{fig:efficiency}).
Refer to the original paper~\cite{ye2024differential} for details.

\begin{figure}[t]
    \centering
    % Subfigure (a)
    \begin{subfigure}[b]{0.22\textwidth}
        \centering
        \includegraphics[width=\textwidth]{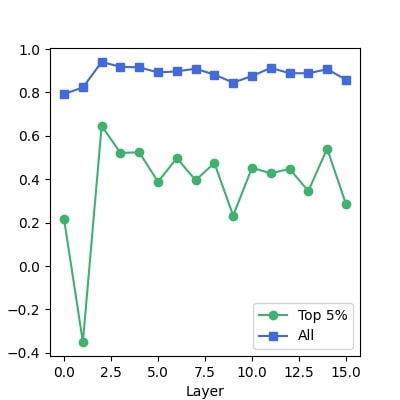}
        \caption{Rank Correlation}
        \label{fig:a}
    \end{subfigure}
    \hspace{0.2em}
    % Subfigure (b)
    \begin{subfigure}[b]{0.22\textwidth}
        \centering
        \includegraphics[width=\textwidth]{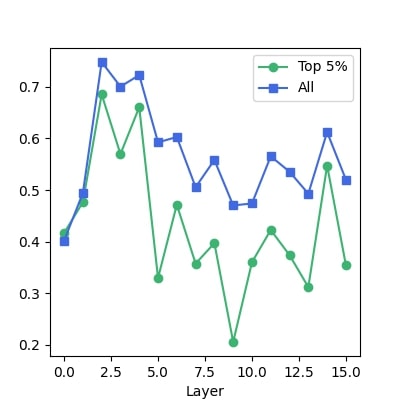}
        \caption{Pearson's Correlation}
        \label{fig:b}
    \end{subfigure}
    \hspace{0.2em}
    % Subfigure (c)
    \begin{subfigure}[b]{0.22\textwidth}
        \centering
        \includegraphics[width=\textwidth]{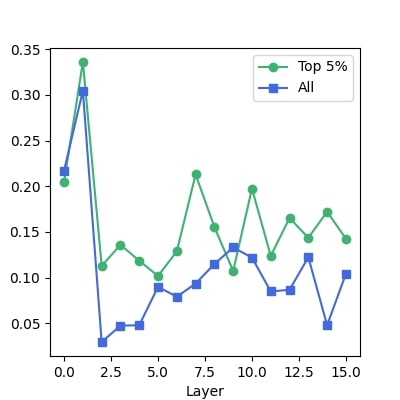}
        \caption{JS Divergence}
        \label{fig:c}
    \end{subfigure}
    \hspace{0.2em}
    % Subfigure (d)
    \begin{subfigure}[b]{0.22\textwidth}
        \centering
        \includegraphics[width=\textwidth]{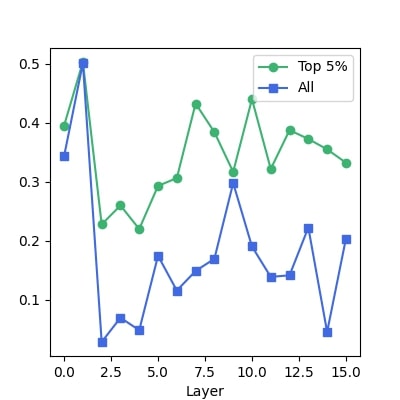}
        \caption{Cosine Distance}
        \label{fig:d}
    \end{subfigure}

    \caption{Attention score comparison between the two groups in \abbr~attention. \textit{Top 5\%} refers to top-5\% tokens with highest attention score in each sequence. It clearly shows that the overlap between the two attention scores is much greater in non-salient tokens. }
    \label{fig:corr}
    \vspace{-1em}
\end{figure}

\subsection{Higher Expressivity via Negative Attentions}
\label{sec:2.2}

The empirical success of \diff~is often attributed to its \textit{noise-canceling} effect, achieved through subtraction between attention groups. Such noise cancellation is commonly hypothesized to enhance performance by inducing sparsity~\cite{ye2024differential}, concentrating attention on relevant context while suppressing irrelevant information. We investigate whether \abbr~attention operates primarily through this lens of conventional sparsity~\cite{tewarson1973sparse,hoefler2021sparsity}.

% This can be an empirical evidence of common-mode noise canceling of \abbr~attention.

\begin{figure}[t]
    \centering
    \begin{subfigure}[b]{0.23\textwidth}
        \centering
        \includegraphics[width=\textwidth]{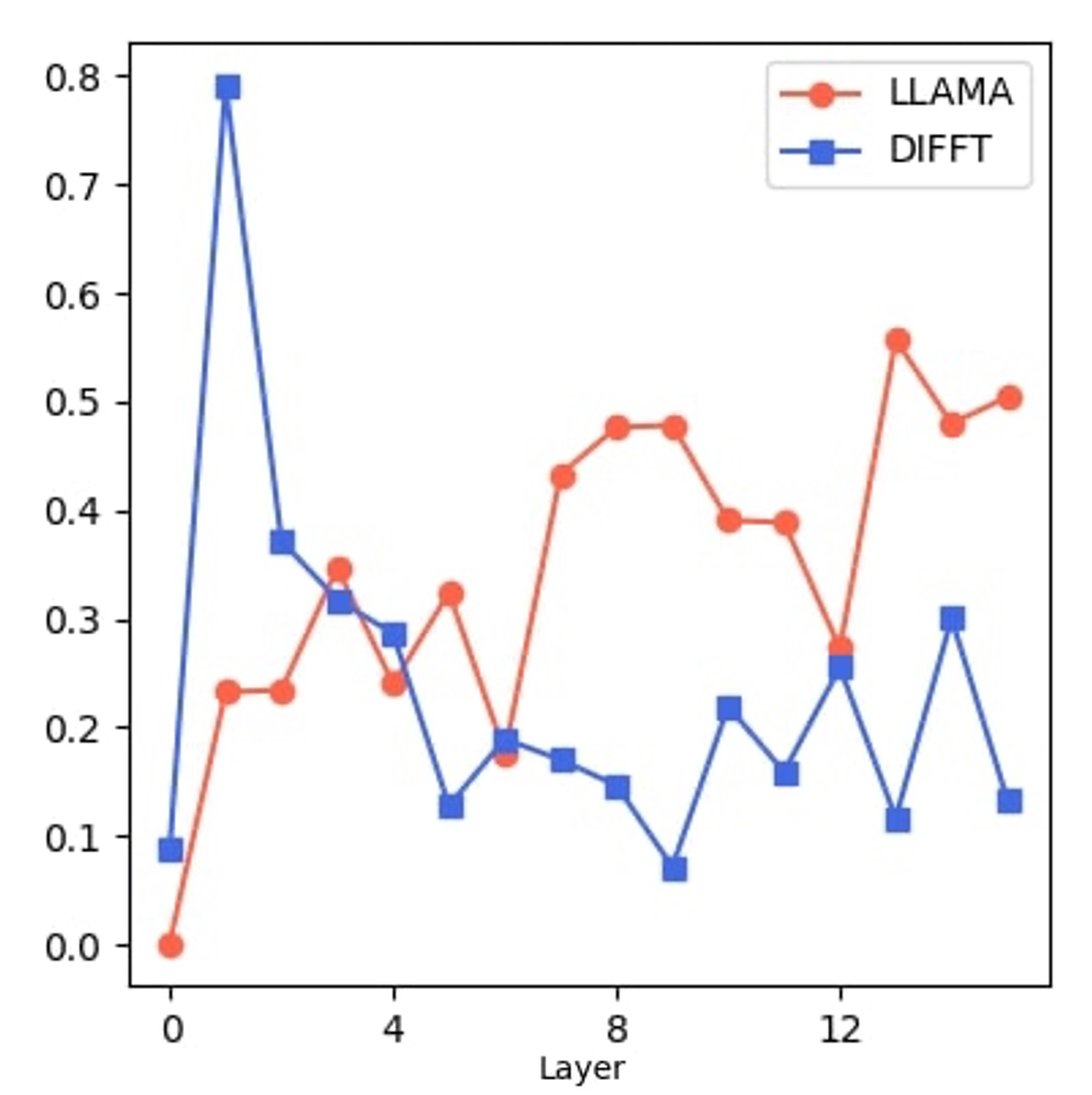}
        \captionsetup{font=footnotesize} % or scriptsize, footnotesize, etc.
        \caption{Sparsity ratio ($\epsilon$=1e-4)}
        \label{fig:sparsity:a}
    \end{subfigure}
    \hspace{0.2em}
    \begin{subfigure}[b]{0.23\textwidth}
        \centering
        \includegraphics[width=\textwidth]{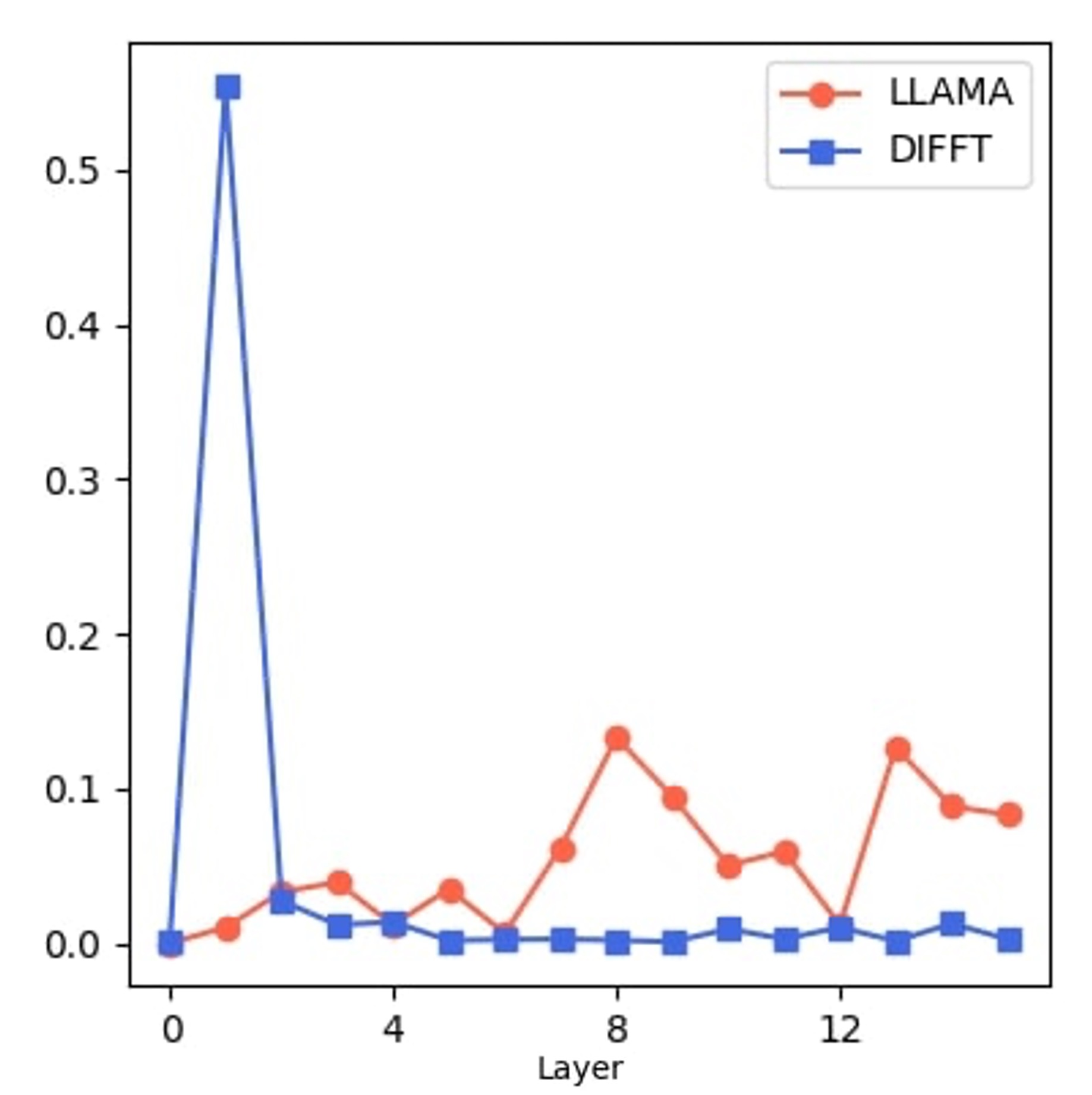}
        \captionsetup{font=footnotesize} % or scriptsize, footnotesize, etc.
        \caption{Sparsity ratio ($\epsilon$=1e-6)}
        \label{fig:sparsity:b}
    \end{subfigure}
    \hspace{0.2em}
    \begin{subfigure}[b]{0.23\textwidth}
        \centering
        \includegraphics[width=\textwidth]{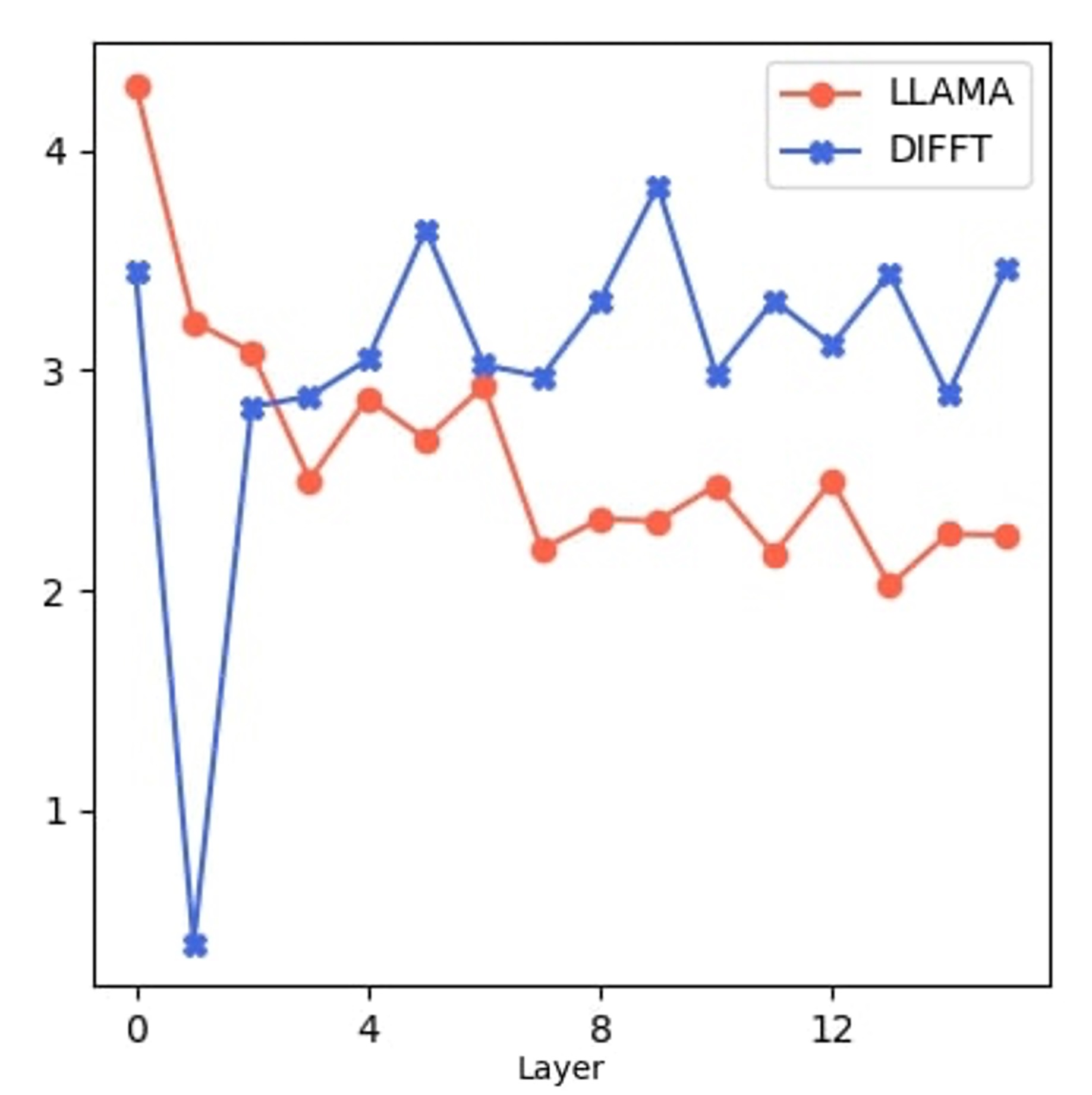}
        \captionsetup{font=footnotesize} % or scriptsize, footnotesize, etc.
        \caption{Attention ent. (abs)}
        \label{fig:sparsity:c}
    \end{subfigure}
    \hspace{0.2em}
    \begin{subfigure}[b]{0.23\textwidth}
        \centering
        \includegraphics[width=\textwidth]{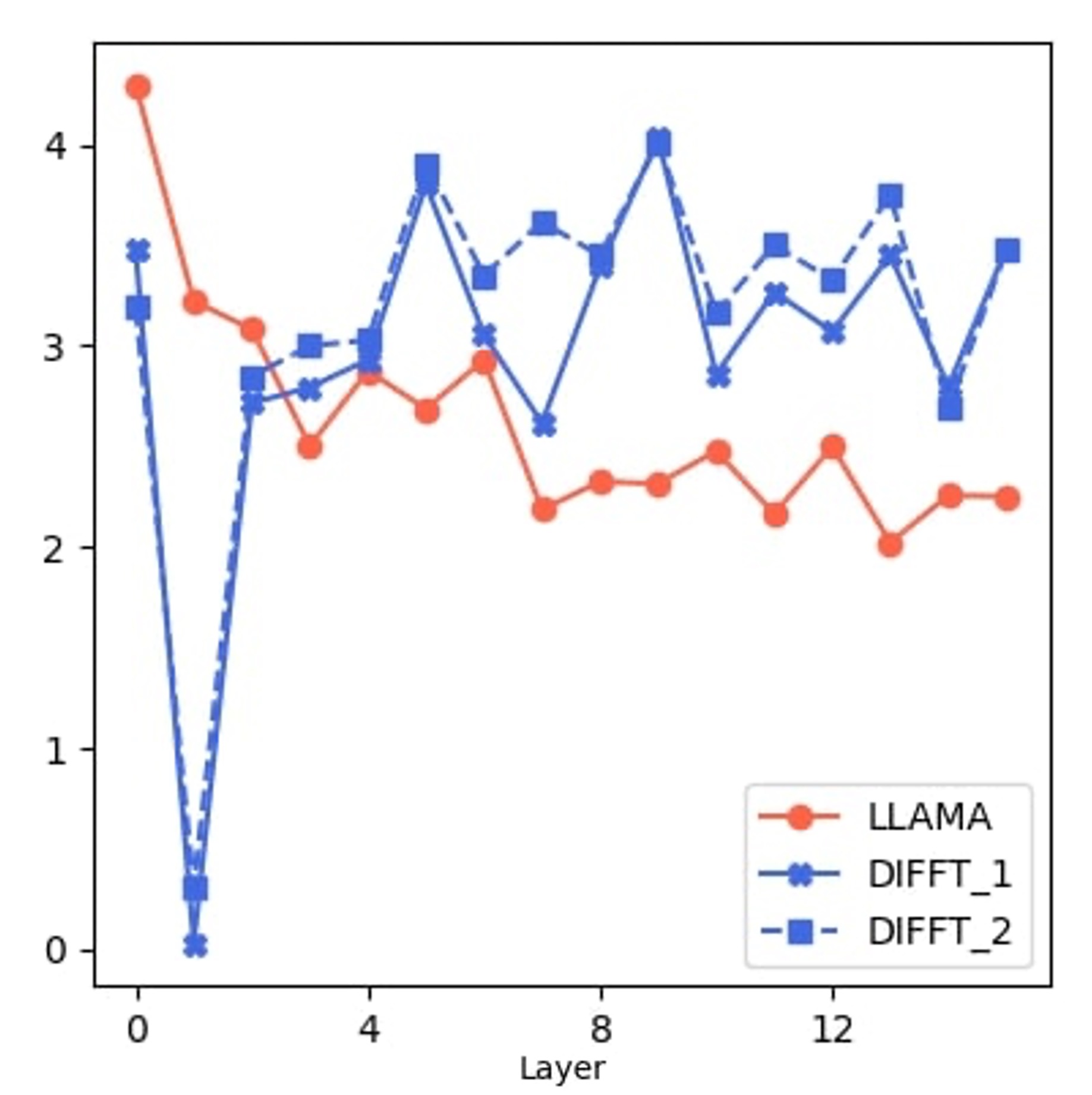}
        \captionsetup{font=footnotesize} % or scriptsize, footnotesize, etc.
        \caption{Attention ent. (group)}
        \label{fig:sparsity:d}
    \end{subfigure}

    \caption{(a), (b): ratio of attention scores whose absolute value is smaller than $\epsilon$. Except for the bottom layers, \diff~displays lower sparsity ratio. (c), (d): Attention score entropy. Entropy in (c) measures magnitude concentration, calculated on renormalized absolute values of the final differential attention scores. Group refers to the two separate attentions in \abbr.}
    \label{fig:sparsity}
    \vspace{-1em}
\end{figure}

% If \abbr~attention indeed cancels noise, one might naturally expect increased sparsity compared to softmax attention, assuming irrelevant context is suppressed. Interestingly, our analysis contradicts this expectation. 
% Fig.\ref{fig:sparsity}(a) and (b) illustrate sparsity ratios
% showing that \abbr~attention actually exhibits lower sparsity except for a few initial layers. Similarly, Fig.\ref{fig:sparsity}(c) and (d) show higher entropy values for \abbr~attention, indicative of less sparsity. Thus, following the conventional definition of sparsity~\cite{tewarson1973sparse,hoefler2021sparsity}, there is little empirical evidence to suggest that \abbr~attention is inherently sparser.
Our analysis of \abbr~attention's dual attention groups (Fig.\ref{fig:corr}) indeed indicates a form of selective filtering. Metrics such as correlations, Jensen-Shannon divergence~\cite{lin1991divergence}, and cosine distance between the groups' attention scores (computed pairwise between corresponding heads) reveal high overall similarity ({\color{royalblue}blue}) but notably weaker correspondence for the most salient tokens ({\color{mediumseagreen}green}).
This points to a selective cancellation where shared, less critical attention patterns are offset by the subtraction, while distinct signals for key tokens are largely preserved or emphasized.
% , suggesting a selective cancellation of shared noise. 
However, this observed filtering does not directly translate to increased sparsity in its traditional definition (\textit{i.e.,} having many close-to-zero values). In fact, Fig.\ref{fig:sparsity}(a) and (b) show that \abbr~attention often exhibits lower sparsity ratios, while Fig.\ref{fig:sparsity}(c) and (d) reveal higher entropy values, both indicative of lower sparsity when compared to standard softmax attention.
% —the proportion of attention scores whose absolute values fall below threshold $\mathbf{\epsilon}$—

% \begin{wrapfigure}{r}{0.44\textwidth}
% \vspace{-1.2em}
% \centering
% \setlength\intextsep{0pt}
% \begin{subfigure}[b]{0.2099\textwidth} % Half of 0.42
%     \centering
%     \includegraphics[width=\linewidth]{figs/negativity/ratio2.jpg}
%     \vspace{-1.7em}
%     \caption{Proportion}
%     \label{fig:negativity:a}
% \end{subfigure}%
% \hfill
% \begin{subfigure}[b]{0.2099\textwidth}
%     \centering
%     \includegraphics[width=\linewidth]{figs/negativity/mean2.jpg}
%     \vspace{-1.7em}
%     \caption{Magnitude}
%     \label{fig:negativity:b}
% \end{subfigure}
% \vspace{-0.3em}
% \caption{The proportion and magnitude of positive/negative attention scores.}
% \vspace{-1.2em}
% \label{fig:negativity}
% \end{wrapfigure}

%%% little faith

\begin{wrapfigure}{r}{0.46\textwidth} % The single wrapfigure environment
    \centering % Center content within the wrapfigure block
    \vspace{-3em} % Your initial vspace

    % --- Start of first figure content (fig:negativity) ---
    \begin{minipage}{\linewidth} % Minipage to group the first figure
        \centering

        % \captionof{figure}{The proportion and magnitude of positive/negative attention scores.} % Use \captionof for figures in minipages
        % \label{fig:negativity}

        \captionsetup{labelformat=empty, textformat=empty} % Makes the caption content itself invisible
        \captionof{figure}{}\label{fig:negativity} 
        \captionsetup{labelformat=default, textformat=default} % Reset to default for subcaptions
        
        \begin{subfigure}[b]{0.49\linewidth} % Use relative width to the minipage/wrapfigure
            \centering
            \includegraphics[width=\linewidth]{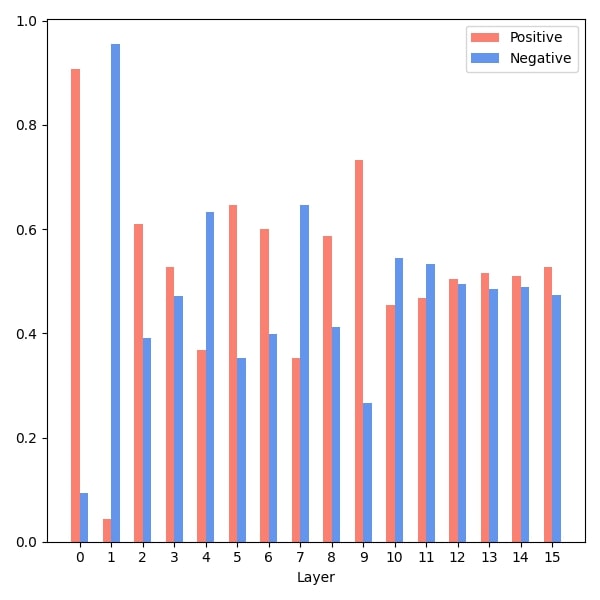}
            \vspace{-1.7em}
            \caption{Proportion}
            \label{fig:negativity:a}
        \end{subfigure}%
        \hfill
        \begin{subfigure}[b]{0.49\linewidth} % Use relative width
            \centering
            \includegraphics[width=\linewidth]{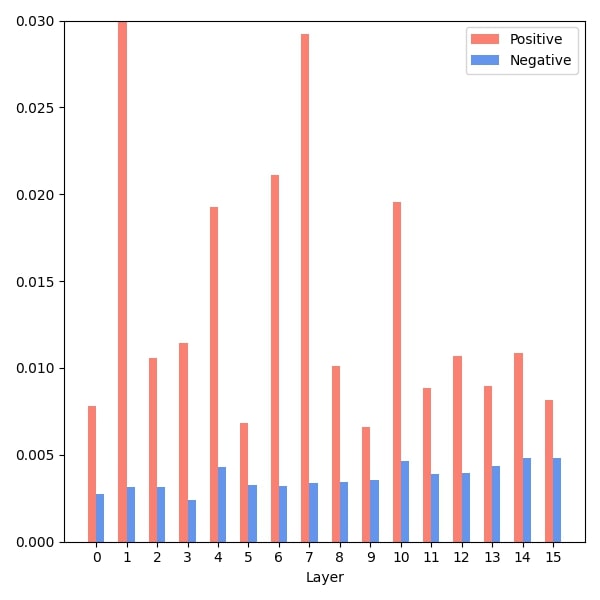}
            \vspace{-1.7em}
            \caption{Magnitude}
            \label{fig:negativity:b}
        \end{subfigure}
        \par 
        \vspace{-0.5em}
        \begin{minipage}{\linewidth} % Create a new minipage for the caption text itself
                                     % This minipage will be centered by the parent's \centering,
                                     % but text *inside* it can be left-aligned.
            \raggedright % Command for left-aligning text within this block
            \footnotesize % Set the font size
            Figure~\ref{fig:negativity}: The proportion and average magnitude of positive/negative attention scores.
        \end{minipage}
    \end{minipage}
    % --- End of first figure content ---

    % \vspace{-0.2em} 

    % --- Start of second figure content (fig:qual) ---
    \begin{minipage}{\linewidth} % Minipage to group the second figure
        \centering
        % Original way for fig:qual (if it's just one image)
        \includegraphics[width=\textwidth]{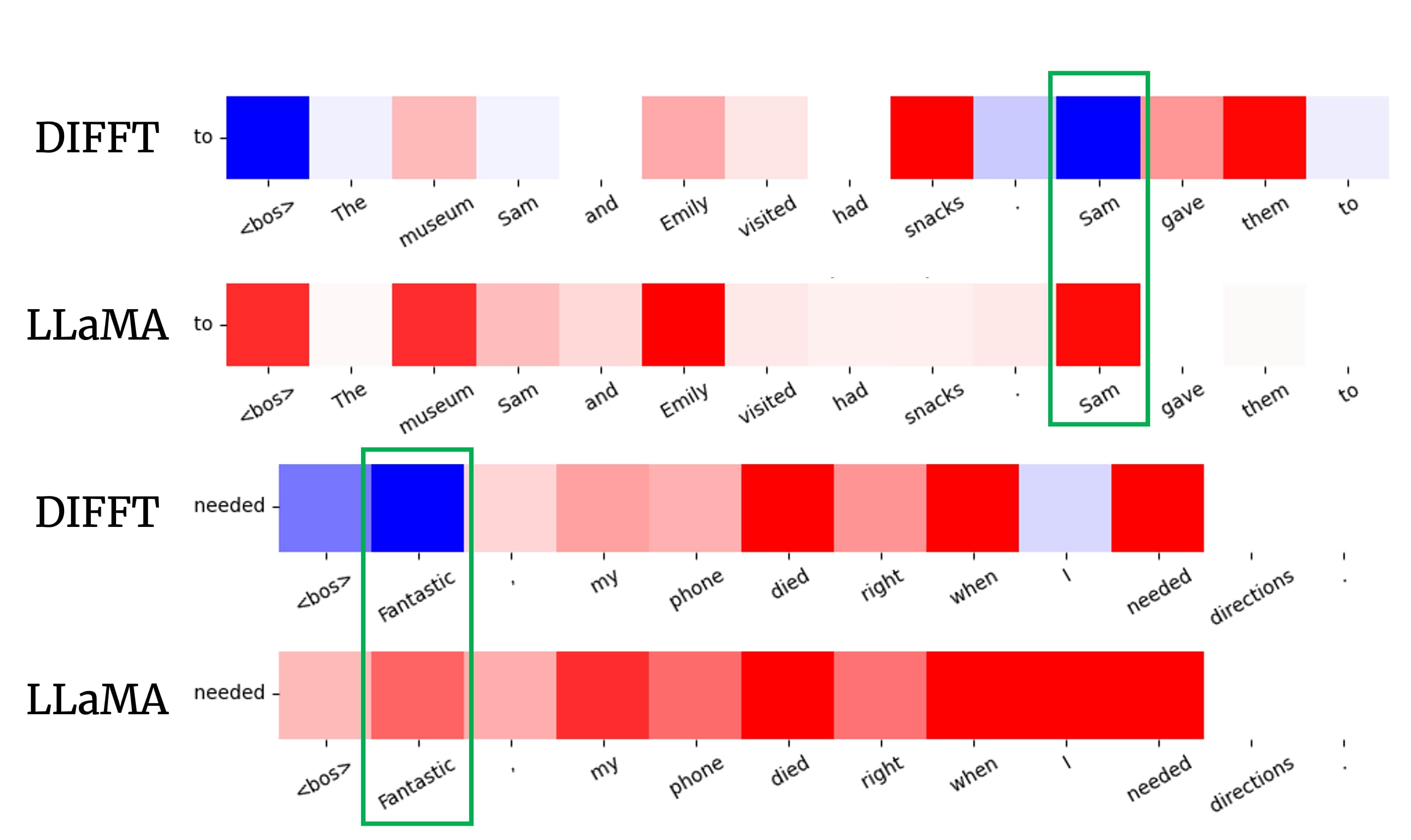} % Or \linewidth if you want it full width of the minipage
        % If you used resizebox:
        % \resizebox{\linewidth}{!}{\includegraphics[width=\textwidth]{figs/qual/qual.jpg}}
        \captionsetup{font=footnotesize} % Apply if needed
        \captionof{figure}{Attention scores on Indirect Object Identification (IOI, top two) and sarcastic expression (bottom two). {\color{plotblue}Blue} indicates negative and {\color{plotred}red} represents positive. Green boxes highlight the difference.
        % \abbr~attention is capable of assigning negative scores on irrelevant context, giving larger flexibility to its representations.
        }
        \label{fig:qual}
    \end{minipage}
    % --- End of second figure content ---

    \vspace{-1.7em} % Your final vspace
\end{wrapfigure}

%%% little faith

% \begin{wrapfigure}{r}{0.44\textwidth}
% \vspace{-2em}
% \centering
% \setlength\intextsep{0pt}
% \resizebox{0.44\textwidth}{!}{%
%     \includegraphics[width=\textwidth]{figs/qual/qual.jpg}
% }
% \caption{Attention scores on Indirect Object Identification (IOI, top two) and sarcastic expression (bottom two). {\color{plotblue}Blue} indicates negative and {\color{plotred}red} represents positive. \abbr~attention is capable of assigning negative scores on irrelevant context, giving larger flexibility to its representations.}
% \vspace{-1em}
% \label{fig:qual}
% \end{wrapfigure}

% This suggests that \abbr~attention's effective information filtering encompasses a more nuanced mechanism than simple sparsity induced by zeroing out. We posit that its efficacy stems from the extensive use of \textit{negative attention scores}. \abbr~attention assigns these negative scores to a substantial fraction of context tokens, approaching 50\% in later layers (Fig.\ref{fig:negativity}). 
% Hence, \abbr~attention does not merely zero out irrelevant contexts, but can more flexibly contextualize them using these signed scores.
This suggests that \abbr~attention’s noise canceling embodies a more nuanced mechanism than simply zeroing out non-salient contexts. As Fig.\ref{fig:negativity} shows, \abbr~attention assigns negative scores to a substantial fraction of context tokens, whose relative attention magnitude generally increases in higher layers. Hence, \abbr~attention does not uniformly zero out irrelevant contexts, but is capable of flexibly contextualizing them using these signed scores. 
As \cite{lv2024more} shows, employing negative attention to explicitly model \textit{negative relevance} in the query-key (QK) circuit provides greater flexibility to the output-value (OV) matrix, reducing its need for implicit information filtering and thereby fostering more expressive representations. Qualitative examples in Fig.\ref{fig:qual}, such as down-weighting irrelevant subject in Indirect Object Identification task~\cite{wang2022interpretability} or non-literal interpretation in sarcasm detection, illustrate how \abbr~attention can achieve a more refined information flow using negative attention (green boxes). This contrasts with standard softmax attention that assigns high scores even to these highly irrelevant contexts, whose sign-insensitivity often burdens its OV matrix with implicit information filtering~\cite{lv2024more}. (Additional examples are in Appendix~\ref{supp:2.5}).

\subsection{Reduced Redundancy among Attention Heads}
\label{sec:2.3}

\begin{figure}[t]
    \centering
    \begin{subfigure}[b]{0.41483\textwidth}
        \centering
        \includegraphics[width=\textwidth]{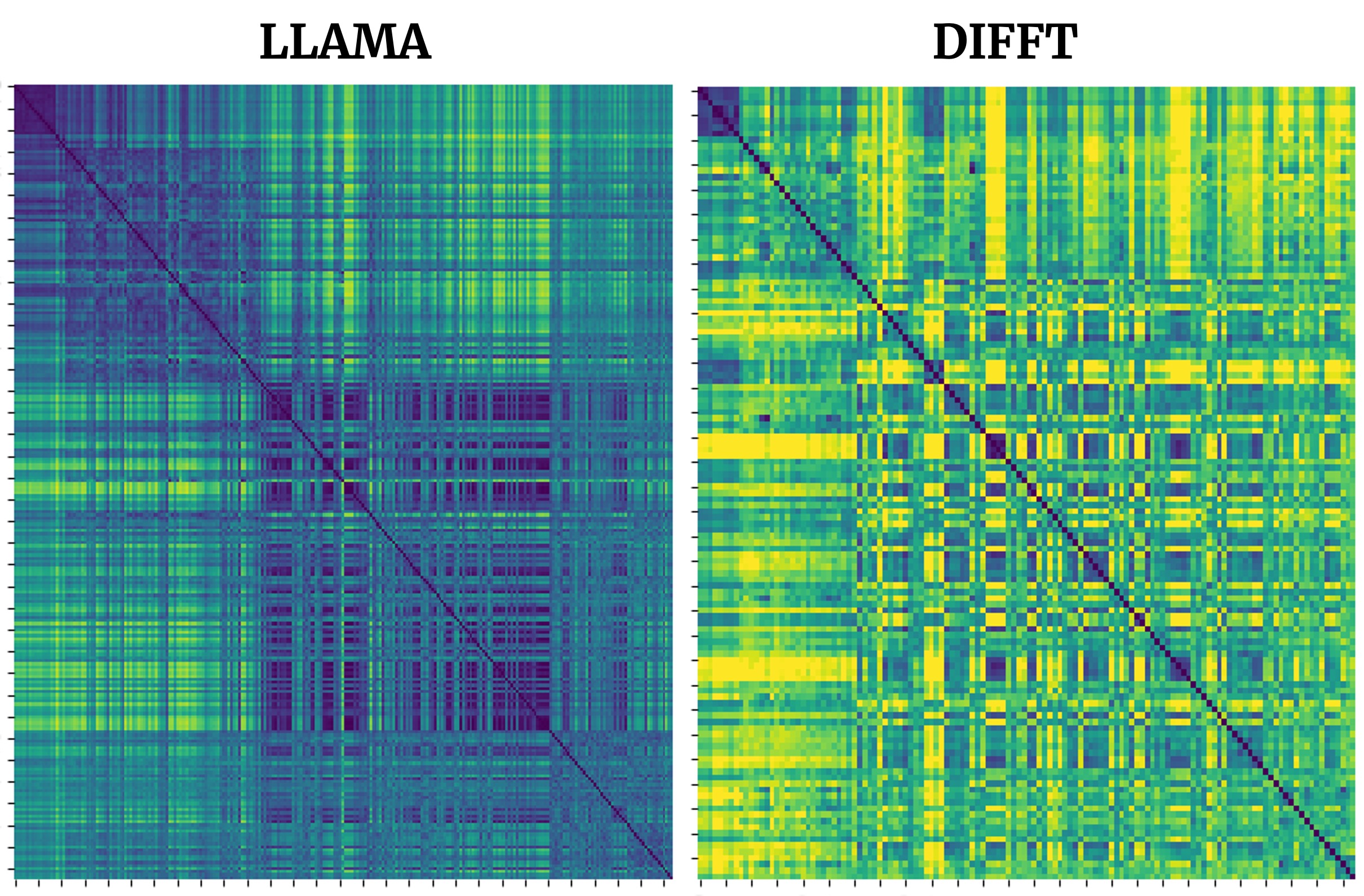}
        \captionsetup{font=footnotesize} % or scriptsize, footnotesize, etc.
        \caption{Cosine distance between attention scores.}
        \label{fig:head_attn:a}
    \end{subfigure}
    % \hfill
    \hspace{1em}
    \begin{subfigure}[b]{0.42561\textwidth}
        \centering
        \includegraphics[width=\textwidth]{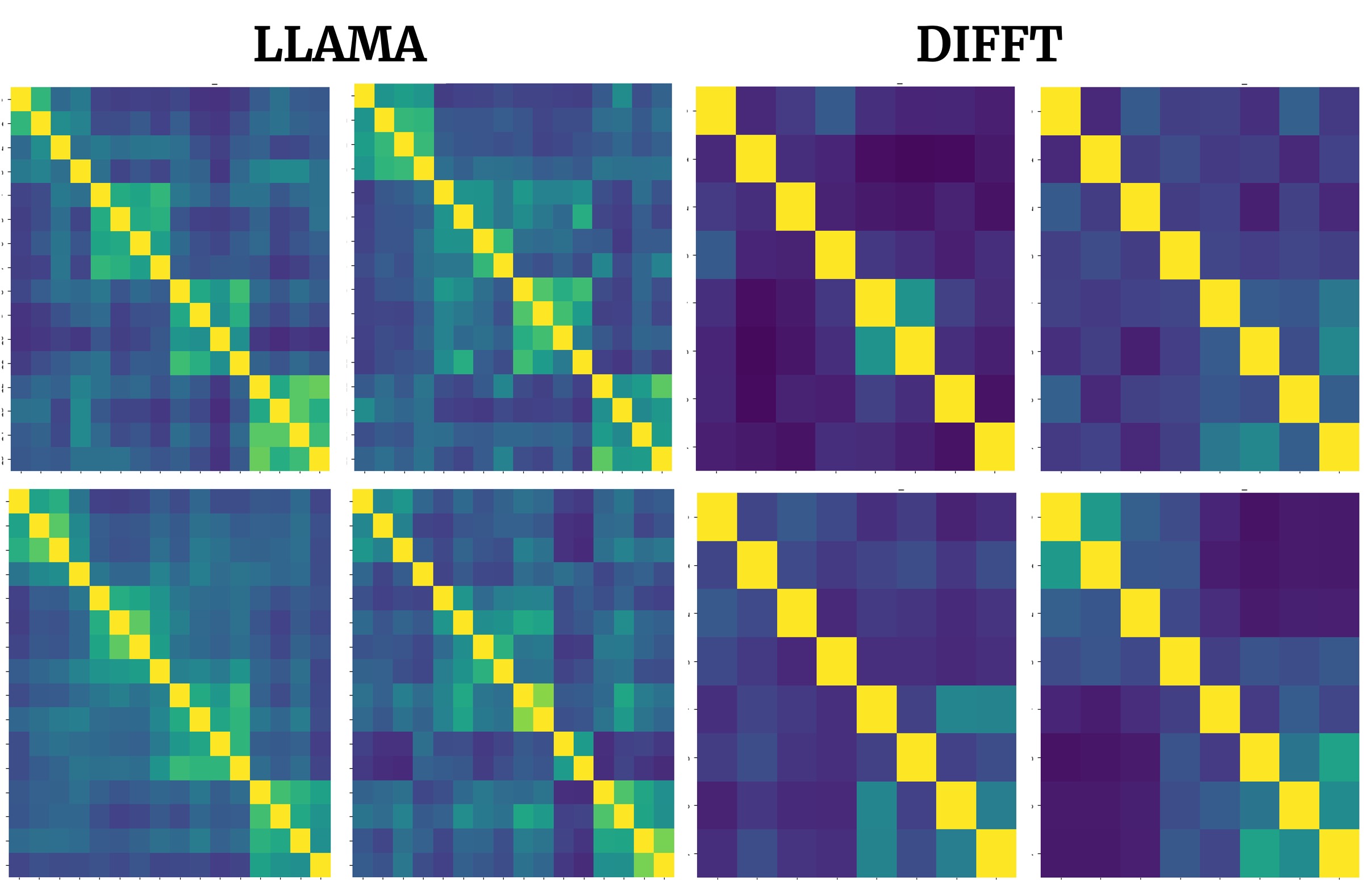}
        \captionsetup{font=footnotesize} % or scriptsize, footnotesize, etc.
        \caption{CKA between attention head features.}
        \label{fig:head_attn:b}
    \end{subfigure}
    \caption{(Left) Pairwise cosine distance between per-head attention scores (flattened across layers) {\color{viridislight}Brighter} indicates larger distance, hence \textbf{\textit{less}} redundancy. (Right) CKA~\cite{nguyen2020wide} between per-head features. {\color{viridislight}Brighter} means higher alignment, hence \textbf{\textit{higher}} redundancy. See Appendix \ref{supp:2.2}.}
    \label{fig:head_attn}
    \vspace{-1em}
\end{figure}

Multi-head self-attention is powerful but can be redundant~\cite{voita2019analyzing,dingpass,li2021differentiable,xia2023sheared,michel2019sixteen,bian2021attention}. Our analysis reveals that \abbr~attention significantly reduces redundancy among attention heads. Fig.\ref{fig:head_attn} presents cosine distance between per-head attention scores (higher distance relates to lower redundancy) and Centered Kernel Alignment~\cite{nguyen2020wide} between value-projected head features (higher alignment translates to higher redundancy). The plots clearly indicate that \abbr~attention exhibits reduced redundancy at both the attention score (left) and feature (right) levels. One might attribute this to \abbr~having fewer effective heads. However, our experiments demonstrate that merely employing fewer, wider attention heads does not alleviate redundancy (see Appendix~\ref{supp:2.2}). We hypothesize that the differential mechanism grants greater flexibility in controlling attention patterns, reducing inter-head redundancy.
% TODO (maybe): add our interpretation as to why this happens. (e.g.,) We hypothesize ...

\begin{wrapfigure}{r}{0.4\textwidth}
\vspace{-1.5em}
\centering
\setlength\intextsep{0pt}
\resizebox{0.35\textwidth}{!}{%
    \includegraphics[width=\textwidth]{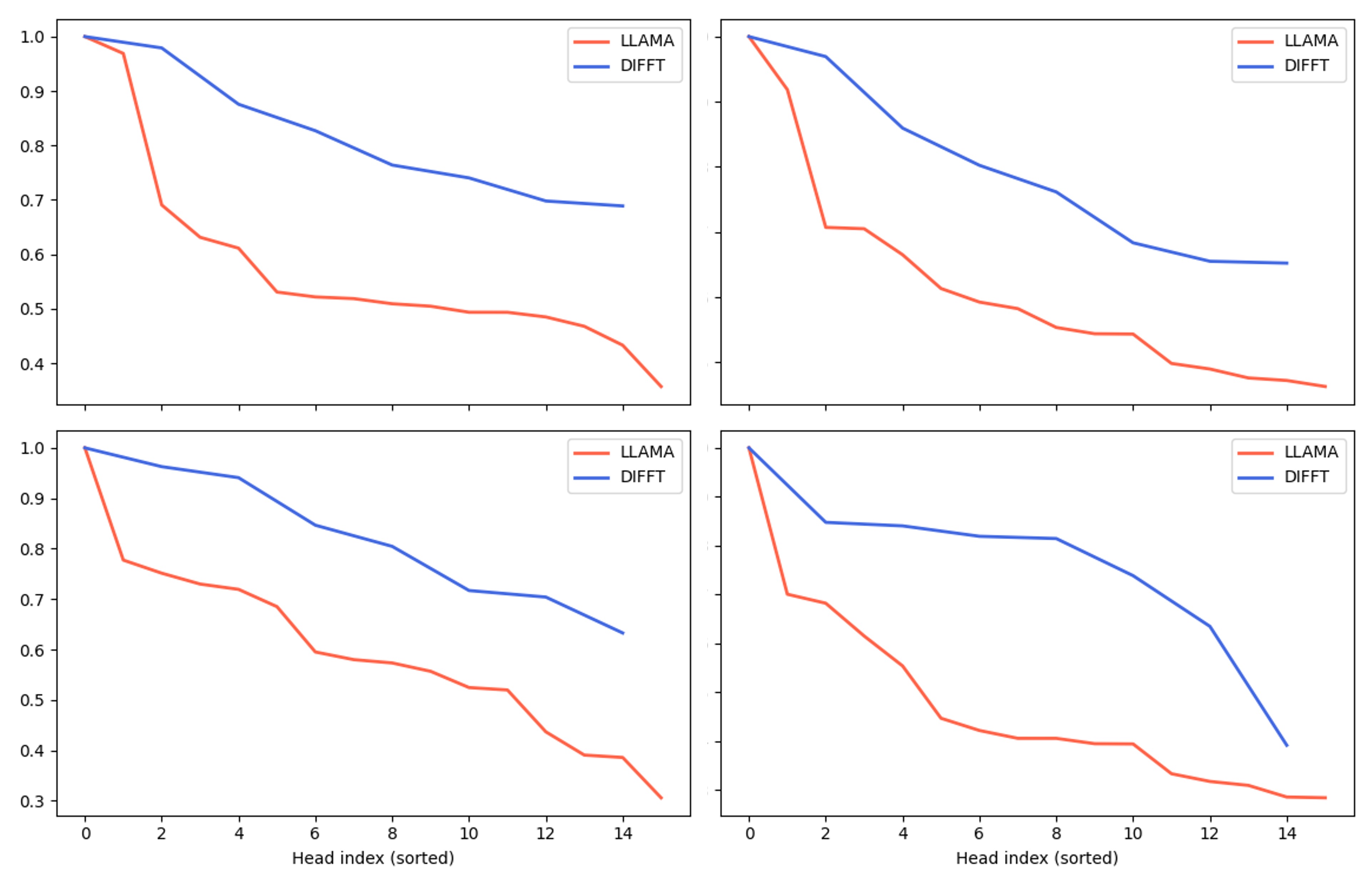}
}
\vspace{-0.5em}
\caption{Layerwise head importance distributions, normalized and sorted.}
\vspace{-2em}
\label{fig:importance}
\end{wrapfigure}

% Examining attention head importance metrics further highlights this reduced redundancy. 
Examining attention head importance provides further insights into head utilization.
Fig.\ref{fig:importance} demonstrates the head importances~\cite{molchanov2019importance,cheng2024mini}, normalized by the maximum value and sorted.
In \diff, importance is distributed more uniformly across attention heads (Fig.\ref{fig:importance} {\color{royalblue}blue}), indicating that each head contributes more evenly to the final representation. Combined with the reduced redundancy, this balanced contribution allows \abbr~attention to capture a broader spectrum of diverse features compared to conventional multi-head attention.

% talk about Group Norm here
\vspace{-0.5em}

\subsection{Improved Learning Dynamics} 
\label{sec:2.4}

% \begin{wrapfigure}{r}{0.4\textwidth}
% \vspace{-2em}
% \centering
% \setlength\intextsep{0pt}
% \resizebox{0.4\textwidth}{!}{%
%     \includegraphics[width=\textwidth]{figs/dynamics/hessian2.jpg}
% }
% \caption{Hessian max eigenvalue spectra at the early (left) and mid stage (right) of training. While normal transformer shows a number of negative Hessian eigenvalues as observed in \cite{park2022vision}, \abbr~has much less. }
% \vspace{-2em}
% \label{fig:hessian}
% \end{wrapfigure}

\begin{figure}[h]
\centering

\begin{minipage}{0.48\linewidth}{
    \includegraphics[width=\linewidth]{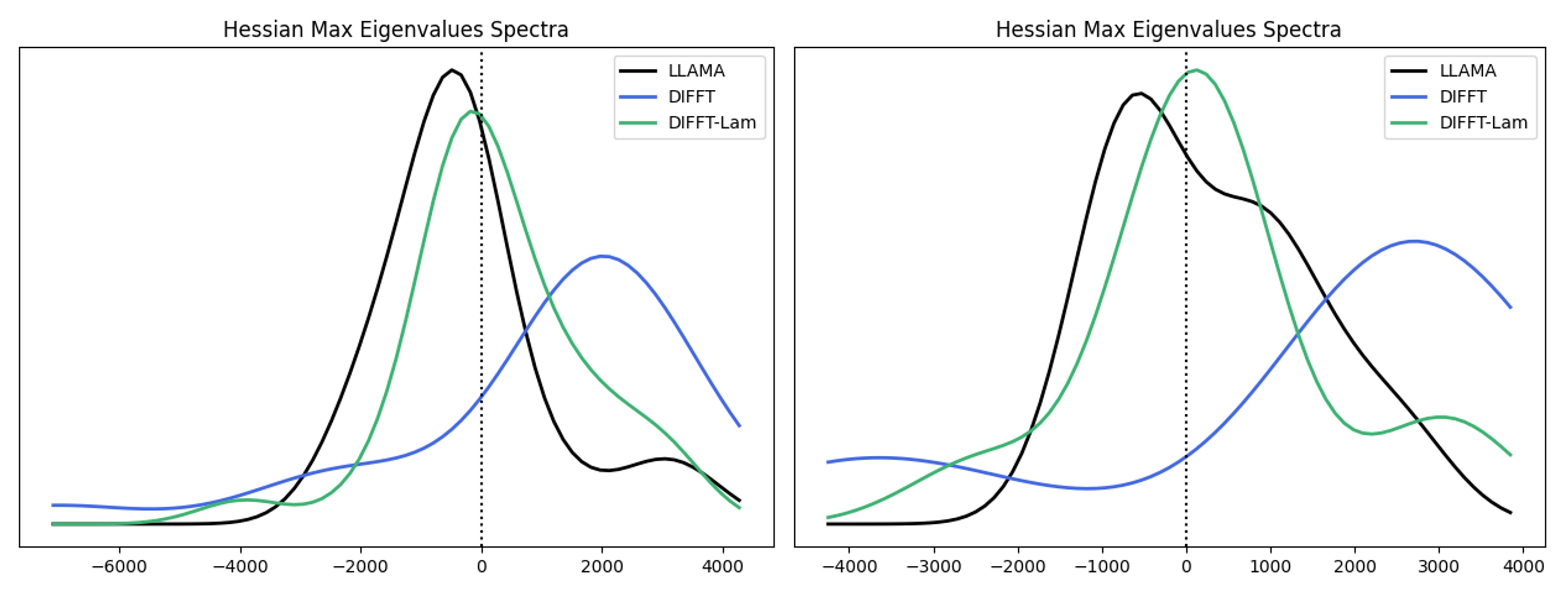}
    \caption{\textbf{Hessian max eigenvalue spectra}. While transformer and \abbr~without learnable $\lambda$ (DIFFT-Lam) shows a number of negative eigenvalues, \abbr~has much less. }
    \label{fig:hessian}
}\end{minipage}
\hfill
\begin{minipage}{0.48\linewidth}{
    \begin{subfigure}[b]{0.49\textwidth}
    \centering
    \includegraphics[width=\linewidth]{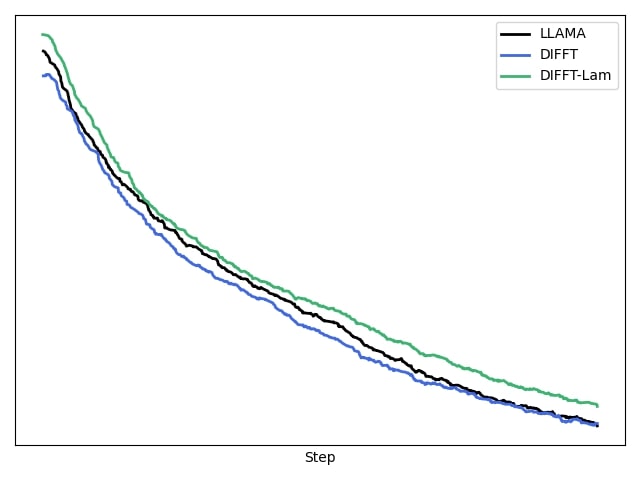}
    \vspace{-1.7em}
    \caption{Loss curve.}
    \label{fig:train:a}
\end{subfigure}%
\hfill
\begin{subfigure}[b]{0.49\textwidth}
    \centering
    \includegraphics[width=\linewidth]{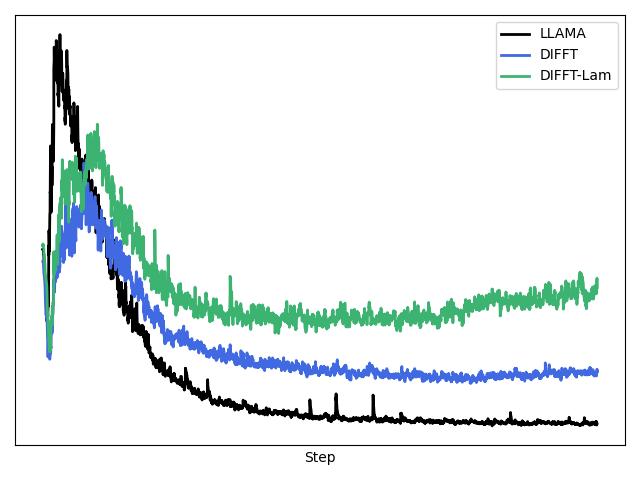}
    \vspace{-1.7em}
    \caption{Gradient norm.}
    \label{fig:train:b}
\end{subfigure}
\vspace{-0.5em}
\caption{\textbf{Loss and gradient norm.} \abbr~shows the best dynamic while DIFFT-Lam, \abbr~with non-learnable $\lambda$, shows instability.}
\label{fig:train}
}\end{minipage}
\end{figure}

% \abbr~attention introduces several novel components including the differential operation and learnable $\lambda$. We further study how the learning dynamics is affected from these changes. Fig.\ref{fig:hessian} shows Hessian maximum eigenvalue spectra of standard transformer and \diff, following the procedure of \cite{park2022vision}. As \cite{park2022vision} discusses, having many negative eigenvalues indicate non-convexity in loss landscape, which can disrupt neural network training particularly in the earlier phase~\cite{park2022blurs,dauphin2014identifying,jastrzebski2021catastrophic,jastrzebski2020break}. We observe much fewer negative eigenvalues in \diff, indicative of better optimization. Notably, this advantage is lost without the learnable $\lambda$ (\textcolor{mediumseagreen}{green}).
\abbr~attention introduces novel components, including the differential operation and a learnable parameter $\lambda$. To understand their impact on learning dynamics, we analyze the Hessian maximum eigenvalue spectra (Fig.\ref{fig:hessian}), following the procedure of~\cite{park2022vision}. As discussed in~\cite{park2022vision}, a high prevalence of negative eigenvalues indicates non-convexity in the loss landscape, which can hinder training, particularly during early phases~\cite{park2022blurs,dauphin2014identifying,jastrzebski2021catastrophic,jastrzebski2020break}. We observe significantly fewer negative eigenvalues for \diff~compared to the standard transformer, suggesting improved optimization dynamics. Notably, this benefit is largely lost when the learnable $\lambda$ is removed (\textcolor{mediumseagreen}{green} line in Fig.\ref{fig:hessian}).

Training statistics further corroborate this finding.
Fig.\ref{fig:train} plots the language modeling loss and gradient norms for the standard and \abbr~transformer. While \abbr~consistently achieves lower loss and more stable grad norms, removing the learnable $\lambda$ notably impairs optimization. We hypothesize that the learnable $\lambda$ plays a key role in stabilizing training dynamics, especially during the early stages.

\section{Differential Extension}
\label{sec:3}

\begin{figure*}[t]
\begin{minipage}{0.38\textwidth}
    \includegraphics[width=0.9\textwidth]{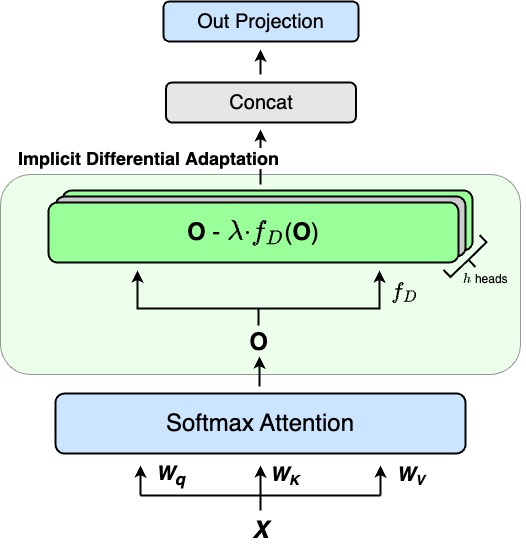}
\end{minipage}
% \hfill
\hspace{0.3em}
\begin{minipage}{0.58\textwidth}
\centering
\begin{lstlisting}[language=python, mathescape, breaklines=true]  
def Attn(X,W_q,W_k,W_v,f_D,$\lambda$,do):
    # standard softmax attention
    Q, K, V = X @ W_q, X @ W_k, X @ W_v
    s = 1 / sqrt(d)
    A = Q @ K.transpose(-1, -2) * s
    O = softmax(A) @ V
    # implicit differential adaptation
    O = O - $\lambda \cdot $f_D(O) if do else O
    return O

def MHA(X,W_q,W_k,W_v,f_D,W_o,$\lambda$,hs):
    O = [Attn(X,...,$\lambda$,do=(i in hs)) for i in range(h)] # hs: selected heads
    return Concat(O) @ W_o
\end{lstlisting}
\end{minipage}
\caption{
\textbf{Differential Extension (\dex)}.
The output value matrix $\mathbf{O}$ is transformed by subtracting a $\lambda$-modulated projection from itself. 
This operation targets a layer-specific subset of attention heads.
}
\vspace{-1em}
\label{fig:method}
\end{figure*}

Based on the insights from Sec.\ref{sec:2}, we present \dex, a framework that integrates differential mechanism into pretrained self-attentions. In designing \dex, we identify three primary desiderata: (1) effectively integrating the beneficial properties of \diff; (2) ensuring a lightweight transition by maximally preserving and leveraging the pretrained knowledge; and (3) minimizing test-time computational or memory overhead. In the following subsections, we describe each component of our framework in detail, explicitly connecting the lessons learned from Sec.\ref{sec:2} to satisfy these desiderata.

\subsection{Implicit Differential Adaptation}
\label{sec:3.1}

% As established in our analysis (Sec.\ref{sec:2.2}), \abbr~attention enhances representational power by enabling the QK circuit to model negative relevance, thereby allowing the OV matrix to focus more on nuanced information processing. 
% While \abbr~attention relies on an explicit subtraction of two attention scores, naively retrofitting this structure onto pretrained model, for instance, by splitting existing heads or duplicating them can lead to either significant loss of pretrained knowledge and instability (former), or introduce prohibitive computational and parameter overhead (latter).
% Our goal with \dex~is to achieve this enhancement in information processing stably and efficiently. 

% Our analysis (Sec.\ref{sec:2.2}) established that \abbr~attention enhances representational power: its QK circuit models negative relevance, enabling the OV matrix to perform more nuanced information processing.
Our analysis (Sec.\ref{sec:2.2}) suggests that \abbr~attention's ability to model negative relevance in its QK circuit enhances representational power by facilitating more nuanced information processing in the OV matrix.
While this is achieved in \abbr~attention by explicitly subtracting two attention scores, naively retrofitting such a dual-group structure onto pretrained models can be problematic. Splitting existing heads into two groups risks significant knowledge loss and instability; duplicating them incurs prohibitive computational and parameter overhead. With \dex, we aim to achieve similar enhancements in information processing, but stably and efficiently.

\dex~introduces its learnable differential mechanism directly to the attention \textit{output} instead of the query-key (QK) circuit, an approach we term \textit{implicit} adaptation. This strategy is motivated by the reusability of pretrained attention magnitude signals, supported by our empirical findings that the \textit{absolute} scores of \abbr~attention often mirror standard softmax scores (Fig.\ref{fig:qual}, Appendix~\ref{supp:2.1}). 
% This design allows \dex~to maximally leverage existing knowledge and maintain efficiency both during training and inference. 
By targeting the OV matrix, which is known to control information flow and perform implicit filtering (Sec.\ref{sec:2}, \cite{lv2024more,elhage2021framework}), \dex~empowers the pretrained attention with improved processing of standard attention patterns.

Formally, our implicit differential adaptation is defined as follows:

\vspace{-0.5em}
\begin{equation}
    \mathbf{O} = \text{softmax}\left(\frac{\mathbf{Q}\mathbf{K}^\top}{\sqrt{d}}\right)\mathbf{V}, \quad \mathbf{O}' = \mathbf{O} - \lambda f_{D}(\mathbf{O}),
    \label{eq:implicit-diff}
\end{equation}

where $f_{D}$ denotes a learnable projection parameterized by $\mathbf{W}_D$ and $\lambda$ is a learnable scalar.
This design offers several notable advantages, including 
lightweight adaptation through effective knowledge reuse,
minimal parameter and test-time compute overhead,
and high compatibility with existing transformers.
We empirically demonstrate that despite being implicit, \dex~effectively delivers the empirical strengths of differential attention.
% minimal computational overhead by directly leveraging pretrained attention patterns, effective knowledge reuse through reinforced OV circuit, and smooth integration via the adaptive scalar $\lambda$. 

\subsection{Selective Adaptation}
\label{sec:3.2}

% Attention heads within multi-head self-attention (MHA) often exhibit functional heterogeneity: some specialize in critical roles, while others may be less contributory or appear redundant~\cite{voita2019analyzing, michel2019sixteen}. 
% Multi-head self-attention (MHA) heads often exhibit functional heterogeneity, ranging from critical specialists to less contributory or redundant ones~\cite{voita2019analyzing, michel2019sixteen}. 
% As Sec.\ref{sec:2.3} shows, attention heads in standard multi-head attention can be highly redundant, and their contribution to the final representation is seldom equal~\cite{voita2019analyzing, michel2019sixteen}.

Attention heads in standard multi-head attention can be highly redundant, and their contribution to the final representation is seldom equal~\cite{voita2019analyzing, michel2019sixteen}. 
% Further motivated by our findings on head utilization (Sec.\ref{sec:2.3}),
% we propose to \textit{leverage} this via selective adaptation,
Further motivated by our findings on effective head utilization in differential attention (Sec.\ref{sec:2.3}), we propose to \textit{leverage} this inherent heterogeneity via selective adaptation,
applying the implicit adaptation (Eq.~\ref{eq:implicit-diff}) only to a subset of heads within each layer, typically targeting those identified as less critical.
% This selective approach aims to enhance potentially underutilized heads without disrupting the specialized functions of essential ones, thereby preserving vital pretrained representations while simultaneously improving the overall representational capacity. We introduce two data-driven head selection strategies:
This selective approach enhances underutilized heads while preserving critical ones, thereby improving overall representational capacity and safeguarding vital pretrained knowledge.
% Our Sec.\ref{sec:2.3} finding that \abbr~attention achieves more diverse and balanced head utilization underscores the common issues of redundancy and unequal contributions in standard MHA~\cite{voita2019analyzing, michel2019sixteen}, motivating our selective adaptation.
% We propose to \textit{leverage} this observed head heterogeneity via selective adaptation, applying the implicit adaptation (Eq.~\ref{eq:implicit-diff}) only to a subset of heads within each layer, typically targeting those identified as less critical. 
% This selective approach enhances underutilized heads while preserving critical ones, thereby improving overall representational capacity and safeguarding vital pretrained knowledge.
We introduce two data-driven head selection strategies:
\vspace{-0.8em}
% \noindent
\paragraph{Low-Importance Head Selection.} The first method selects heads based on low representational importance, following headwise importance criteria established in \cite{molchanov2019importance,cheng2024mini}. We compute importance scores and apply differential adaptation to the top-$k$ heads with the lowest scores in each layer.
\vspace{-0.8em}\
% \noindent
\paragraph{High-Entropy Head Selection.} The second strategy targets attention heads with high entropy, a state often associated with weaker representational focus, reduced functional specialization, or potential under-utilization~\cite{zhang2024attention,jha2025entropy,meng2025polaformer,li2023interpreting}. 
Similarly, we select and adapt the top-$k$ heads demonstrating the highest entropy within each layer.
% Accordingly, we select and adapt the top-$k$ heads demonstrating the highest entropy within each layer.
% targets heads exhibiting high attention entropy, typically indicative of 
% weaker representational focus, lower functional specialization or under-utilization~\cite{zhang2024attention,jha2025entropy,meng2025polaformer,li2023interpreting}. 
% Empirical evidence (Fig.\ref{fig:sparsity:d}) also suggests higher entropy in the attention heads being subtracted within differential attention frameworks. 

\subsection{Balancing Adaptation with Pretrained Knowledge via $\lambda$-Annealing}
\label{sec:3.3}

Our analysis in Sec.\ref{sec:2.4} reveals that adaptive modulation of the differential mechanism is critical for stable optimization.
In our scenario, maintaining a careful balance between pretrained knowledge and newly introduced architectural modifications is crucial.
Zero-initializing the learnable $\lambda$ would be a typical way to safely introduce \dex~\cite{hu2022lora,zhang2023adding,guo2023animatediff}, but that alone does not sufficiently encourage the model to adopt the differential mechanism, as $\lambda$ could remain near zero if the pretrained model is already strong.
To facilitate a stable and effective transition, we propose a scheduled annealing of $\lambda$:

\vspace{-0.5em}

\begin{equation}
    \lambda(t) = (1 - \alpha) \left[\frac{t}{T}\lambda_{\text{init}}\right] + \alpha \lambda_{\text{learn}}, 
    \quad \alpha = \min\left(1, \frac{t}{T}\right)
\end{equation}

where $t$ is the current training step, $T$ is the annealing duration, $\lambda_{\text{init}}$ is a constant,
% adopted from \diff~\cite{ye2024differential}, 
and $\lambda_{\text{learn}}$ is a learnable parameter initialized around zero. 
This schedule initiates $\lambda(t)$ with zero for stability, uses annealed $\lambda_{\text{init}}$ to provide a gradual learning signal for the differential mechanism (e.g., $\mathbf{W}_D$) when $0<t<T$, and transitions control to the learnable $\lambda_{\text{learn}}$ for optimal adaptation as $t \ge T$.
% In essence, we interpolate between an annealed reference and the fully learnable parameter, smoothly guiding the transition and preventing $\lambda$ from stagnating near zero.
% This design prevents premature divergence from pretrained knowledge while naturally promoting effective adaptation as training progresses. 
% We note that partial realization of $\lambda_{\text{init}}$ poses no issue due to our setting differing from that of~\cite{ye2024differential}; empirical experiments confirm minimal sensitivity to the specific choice of $\lambda_{\text{init}}$ (see Appendix \ref{}).
% \begin{equation}
%     \lambda(t) = (1 - \alpha) \left[\alpha\lambda_{\text{init}}\right] + \alpha \lambda_{\text{learn}}, 
%     \quad \alpha = \min\left(1, \frac{t}{T}\right)
% \end{equation}
% Basically, we interpolate between $\alpha\lambda_{init}$ and $\lambda_{learn}$, where the former is annealed to serve as a gentle pacemaker for the latter.
% We note that $\lambda_{\text{init}}$ not being fully realized poses no issue due to our differing setting from~\cite{ye2024differential}; empirical experiments also demonstrate minimal sensitivity to the specific choice of $\lambda_{\text{init}}$.

\subsection{Overall Framework}
\label{sec:3.4}

The complete formulation of \dex~for a given head $h$ is expressed as follows:

\vspace{-0.5em}

\begin{equation}
    \mathbf{O} = \text{softmax}\left(\frac{\mathbf{Q}\mathbf{K}^\top}{\sqrt{d}}\right)\mathbf{V}, \quad \mathbf{O}' = \mathbf{O} - \lambda(t)\mathbb{I}(h \in \mathcal{H}) f_{D}(\mathbf{O}),
    \label{eq:overall-framework}
\end{equation}

where $\mathcal{H}$ is the set of heads selected for differential adaptation, and $\mathbf{O}'$ is concatenated across all heads and passed into the output projection. During training, we update $\mathbf{W_K},\mathbf{W_V}$, and $\mathbf{W_O}$ along with $\mathbf{W_D}$ and $\lambda_{learn}$ within self-attention, keeping all other parameters (\textit{e.g.,} FFN) frozen. 
This targeted update strategy provides the necessary flexibility to integrate \dex~into standard transformers, while keeping the training lightweight, especially under standard GQA~\cite{ainslie2023gqa} setting.

% move to appendix
% While the linear \dex~operation could theoretically be learned implicitly by fine-tuning $W_V$ or $W_O$, our explicit parameterization using $W_D$ offers practical benefits. It enables direct control through $\lambda$-annealing (Sec.\ref{sec:3.3}) and provides useful inductive bias for learning the differential filtering task, potentially aiding optimization. 
% This explicit factorization complements standard fine-tuning, leading to improved empirical results (Sec.\ref{sec:4}).

\section{Experiments}
\label{sec:4}

\diff~has demonstrated strong performance across a wide variety of tasks, including general language modeling, key information retrieval, and in-context learning.
We quantitatively validate the effectiveness of \dex~in integrating these strengths into pretrained LLMs. 
We conduct ablation experiments and analyses to further verify our design choices.
% We further study if a pretrained transformer, after \dex, attains core mechanisms of \diff~illustrated in Sec.\ref{sec:2}. Lastly, we conduct extensive ablation experiments to verify our design choices.
% study whether a pretrained transformer, after applying \dex, attains the core mechanisms of \abbr~illustrated in Sec.\ref{sec:2}. Lastly, we 

\subsection{Language Modeling Evaluation}
\label{sec:4.1}

\begin{table}[t]
\centering
\caption{Language modeling benchmark scores across model variants and training methods. \colorbox{win}{Green} indicates improvement over the baseline, while \colorbox{lose}{gray} indicates a decrease. 
}
\label{tab:main}
\resizebox{\textwidth}{!}{%
\begin{tabular}{lccccccccccc|cc}
\toprule
\textbf{Model} & \textbf{Arc-C} & \textbf{Arc-E} & \textbf{BoolQ} & \textbf{COPA} & \textbf{Hellaswag} & \textbf{MNLI} & \textbf{OBQA} & \textbf{PIQA} & \textbf{WIC} & \textbf{Winogrande} & \textbf{WSC} & \textbf{AVG} & \textbf{$\Delta$} \\
\regmidrule
% --- Llama-3.2 3B ---
\\[-0.8em]
\textbf{\textit{Llama-3B}}       & 46.3 & 71.7 & 73.1 & 85.0 & 73.6 & 35.0 & 43.2 & 77.5 & 49.8 & 69.1 & 37.5 & 60.2 & - \\
\thinmidrule
\\[-0.8em]
LoRA (r=8)     & \cellcolor{lose}43.4 & \cellcolor{lose}70.2 & \cellcolor{win}75.3 & \cellcolor{lose}82.0 & \cellcolor{win}74.2 & \cellcolor{win}54.5 & \cellcolor{lose}43.0 & \cellcolor{lose}77.1 & \cellcolor{win}53.8 & \cellcolor{win}70.1 & \cellcolor{lose}36.5 & \cellcolor{win}61.8 & \textcolor{deltaGreen}{+1.6} \\
LoRA (r=32)    & \cellcolor{lose}43.7 & \cellcolor{win}72.0 & \cellcolor{win}76.2 & \cellcolor{lose}83.0 & \cellcolor{win}74.7 & \cellcolor{win}46.7 & \cellcolor{lose}43.2 & \cellcolor{win}77.7 & \cellcolor{win}55.2 & \cellcolor{win}70.0 & \cellcolor{lose}36.5 & \cellcolor{win}61.7 & \textcolor{deltaGreen}{+1.5} \\
PiSSA          & \cellcolor{lose}45.4 & \cellcolor{win}73.8 & \cellcolor{win}74.1 & \cellcolor{lose}82.0 & \cellcolor{win}74.3 & \cellcolor{win}46.6 & \cellcolor{lose}42.4 & \cellcolor{win}78.3 & \cellcolor{win}56.1 & \cellcolor{win}69.9 & \cellcolor{win}38.5 & \cellcolor{win}61.9 & \textcolor{deltaGreen}{+1.7} \\
\thinmidrule
\\[-0.8em]
FT             & \cellcolor{lose}45.7 & \cellcolor{win}73.7 & \cellcolor{win}73.8 & \cellcolor{lose}84.0 & \cellcolor{win}74.7 & \cellcolor{win}38.5 & \cellcolor{lose}41.4 & \cellcolor{win}78.0 & \cellcolor{win}55.3 & \cellcolor{win}70.7 & \cellcolor{win}40.4 & \cellcolor{win}61.5 & \textcolor{deltaGreen}{+1.3} \\
GaLore         & \cellcolor{lose}46.1 & \cellcolor{win}74.9 & \cellcolor{win}76.2 & \cellcolor{win}87.0 & \cellcolor{win}74.1 & \cellcolor{lose}33.1 & \cellcolor{lose}42.6 & \cellcolor{win}77.9 & \cellcolor{win}53.0 & \cellcolor{win}70.2 & \cellcolor{win}38.5 & \cellcolor{win}61.2 & \textcolor{deltaGreen}{+1.0} \\
APOLLO         & \cellcolor{lose}45.8 & \cellcolor{win}74.4 & \cellcolor{win}73.5 & \cellcolor{lose}84.0 & \cellcolor{win}74.7 & \cellcolor{win}35.0 & \cellcolor{lose}42.8 & \cellcolor{lose}77.5 & \cellcolor{win}56.1 & \cellcolor{win}70.2 & \cellcolor{win}45.2 & \cellcolor{win}61.7 & \textcolor{deltaGreen}{+1.5} \\ % APOLLO AVG was 61.7 in image, Delta +1.5
\thinmidrule
\\[-0.8em]
Ours           & \cellcolor{lose}45.5 & \cellcolor{win}73.3 & \cellcolor{win}74.8 & \cellcolor{lose}84.0 & \cellcolor{win}74.1 & \cellcolor{win}49.5 & \cellcolor{lose}42.6 & \cellcolor{win}78.2 & \cellcolor{win}51.9 & \cellcolor{win}69.1 & \cellcolor{win}63.5 & \cellcolor{win}\textbf{64.2} & \textcolor{deltaGreen}{\textbf{+4.0}} \\
\regmidrule
% --- Llama-3.2 1B ---
\\[-0.8em]
\textbf{\textit{Llama-1B}}       & 36.3 & 60.6 & 63.4 & 77.0 & 63.6 & 36.0 & 37.2 & 74.5 & 48.6 & 59.9 & 42.3 & 54.5 & - \\
\thinmidrule
\\[-0.8em]
LoRA (r=8)     & \cellcolor{lose}34.6 & \cellcolor{win}63.3 & \cellcolor{lose}46.4 & \cellcolor{win}78.0 & \cellcolor{win}64.1 & \cellcolor{lose}32.9 & \cellcolor{lose}36.6 & \cellcolor{win}75.1 & \cellcolor{lose}47.9 & \cellcolor{win}60.9 & \cellcolor{lose}40.4 & \cellcolor{lose}52.7 & \textcolor{deltaRed}{-1.8} \\
LoRA (r=32)    & \cellcolor{lose}35.9 & \cellcolor{win}65.4 & \cellcolor{lose}61.5 & \cellcolor{win}78.0 & \cellcolor{win}64.4 & \cellcolor{lose}32.6 & \cellcolor{win}38.2 & \cellcolor{win}75.1 & \cellcolor{win}48.7 & \cellcolor{win}60.3 & \cellcolor{lose}37.5 & \cellcolor{lose}54.3 & \textcolor{deltaRed}{-0.2} \\
PiSSA          & \cellcolor{lose}36.3 & \cellcolor{win}65.2 & \cellcolor{lose}59.8 & \cellcolor{win}79.0 & \cellcolor{win}64.2& \cellcolor{lose}33.1 & \cellcolor{lose}37.2 & \cellcolor{win}75.1 & \cellcolor{win}49.7 & \cellcolor{win}60.7 & \cellcolor{lose}38.5 & \cellcolor{lose}54.4 & \textcolor{deltaRed}{-0.1} \\
\thinmidrule
\\[-0.8em]
FT             & \cellcolor{win}36.8 & \cellcolor{win}65.5 & \cellcolor{lose}60.7 & \cellcolor{lose}76.0 & \cellcolor{win}64.5 & \cellcolor{win}41.2 & \cellcolor{win}38.2 & \cellcolor{win}74.9 & \cellcolor{win}49.2 & \cellcolor{win}60.7 & \cellcolor{lose}36.5 & \cellcolor{win}54.9 & \textcolor{deltaGreen}{+0.4} \\
GaLore         & \cellcolor{lose}36.3 & \cellcolor{win}65.7 & \cellcolor{lose}60.2 & \cellcolor{lose}77.0 & \cellcolor{win}64.2 & \cellcolor{lose}34.4 & \cellcolor{win}37.6 & \cellcolor{win}75.2 & \cellcolor{win}50.6 & \cellcolor{win}60.7 & \cellcolor{lose}36.5 & \cellcolor{lose}54.4 & \textcolor{deltaRed}{-0.1} \\
APOLLO         & \cellcolor{win}37.1 & \cellcolor{win}65.0 & \cellcolor{lose}58.1 & \cellcolor{lose}77.0 & \cellcolor{win}64.4 & \cellcolor{win}37.5 & \cellcolor{lose}36.8 & \cellcolor{win}74.9 & \cellcolor{win}51.6 & \cellcolor{win}60.4 & \cellcolor{lose}36.5 & \cellcolor{win}54.5 & \textcolor{deltaGreen}{+0.0} \\ 
\thinmidrule
\\[-0.8em]
Ours           & \cellcolor{lose}35.2 & \cellcolor{win}64.2 & \cellcolor{lose}57.8 & \cellcolor{win}79.0 & \cellcolor{win}64.0 & \cellcolor{win}38.0 & \cellcolor{win}38.0 & \cellcolor{win}75.0 & \cellcolor{win}51.9 & \cellcolor{win}60.6 & \cellcolor{win}48.1 & \cellcolor{win}\textbf{55.6} & \textcolor{deltaGreen}{\textbf{+1.1}} \\
\regmidrule
% --- Qwen-2.5 1.5B ---
\\[-0.8em]
\textbf{\textit{Qwen-1.5B}}       & 45.1 & 72.2 & 72.8 & 83.0 & 67.8 & 52.6 & 40.6 & 76.0 & 53.0 & 63.5 & 57.7 & 62.2 & - \\
\thinmidrule
\\[-0.8em]
LoRA (r=8)     & \cellcolor{lose}43.3 & \cellcolor{lose}70.3 & \cellcolor{win}73.5 & \cellcolor{win}84.0 & \cellcolor{lose}67.5 & \cellcolor{lose}49.3 & \cellcolor{lose}39.2 & \cellcolor{lose}75.1 & \cellcolor{win}53.3 & \cellcolor{win}64.3 & \cellcolor{lose}51.0 & \cellcolor{lose}61.0 & \textcolor{deltaRed}{-1.2} \\ % AVG is 51.8, same as baseline
LoRA (r=32)    & \cellcolor{lose}43.4 & \cellcolor{lose}70.2 & \cellcolor{lose}71.0 & \cellcolor{win}85.0 & \cellcolor{lose}67.5 & \cellcolor{lose}50.7 & \cellcolor{lose}39.2 & \cellcolor{lose}75.5 & \cellcolor{lose}52.0 & \cellcolor{win}64.7 & \cellcolor{lose}47.1 & \cellcolor{lose}60.6 & \textcolor{deltaRed}{-1.6} \\
PiSSA          & \cellcolor{lose}44.3 & \cellcolor{lose}70.1 & \cellcolor{lose}72.6 & \cellcolor{win}84.0 & \cellcolor{lose}66.7 & \cellcolor{lose}47.5 & \cellcolor{lose}40.0 & \cellcolor{lose}74.3 & \cellcolor{win}54.7 & \cellcolor{win}63.9 & \cellcolor{lose}52.9 & \cellcolor{lose}61.0 & \textcolor{deltaRed}{-1.2} \\
\thinmidrule
\\[-0.8em]
FT             & \cellcolor{lose}43.9 & \cellcolor{lose}71.9 & \cellcolor{lose}68.7 & \cellcolor{win}84.0 & \cellcolor{lose}67.6 & \cellcolor{lose}51.5 & \cellcolor{lose}40.2 & \cellcolor{lose}75.7 & \cellcolor{win}53.6 & \cellcolor{win}64.5 & \cellcolor{lose}48.1 & \cellcolor{lose}60.9 & \textcolor{deltaRed}{-1.3} \\
GaLore         & \cellcolor{lose}44.3 & \cellcolor{win}72.7 & \cellcolor{lose}72.0 & \cellcolor{win}84.0 & \cellcolor{lose}67.4 & \cellcolor{lose}47.6 & \cellcolor{lose}39.6 & \cellcolor{lose}75.0 & \cellcolor{win}53.1 & \cellcolor{win}64.7 & \cellcolor{lose}51.9 & \cellcolor{lose}61.1 & \textcolor{deltaRed}{-1.1} \\ 
APOLLO         & \cellcolor{lose}45.1 & \cellcolor{win}73.4 & \cellcolor{lose}72.4 & \cellcolor{lose}83.0 & \cellcolor{lose}67.7 & \cellcolor{lose}50.1 & \cellcolor{lose}39.4 & \cellcolor{lose}75.7 & \cellcolor{win}53.9 & \cellcolor{win}64.8 & \cellcolor{lose}43.3 & \cellcolor{lose}60.8 & \textcolor{deltaRed}{-1.4} \\ 
\\[-0.8em]
\thinmidrule
\\[-0.8em]
Ours           & \cellcolor{win}45.3 & \cellcolor{win}74.1 & \cellcolor{lose}70.1 & \cellcolor{win}84.0 & \cellcolor{win}67.8 & \cellcolor{lose}50.2 & \cellcolor{win}40.8 & \cellcolor{win}76.4 & \cellcolor{win}53.3 & \cellcolor{lose}63.2 & \cellcolor{win}61.6 & \cellcolor{win}\textbf{62.4} & \textcolor{deltaGreen}{\textbf{+0.2}} \\
\regmidrule
% --- Qwen-2.5 0.5B ---
\\[-0.8em]
\textbf{\textit{Qwen-0.5B}}       & 31.8 & 58.7 & 62.3 & 74.0 & 52.2 & 38.3 & 35.4 & 69.9 & 49.2 & 56.2 & 41.3 & 51.8 & - \\
\thinmidrule
\\[-0.8em]
LoRA (r=8)     & \cellcolor{win}34.3 & \cellcolor{win}66.1 & \cellcolor{lose}57.2 & \cellcolor{lose}74.0 & \cellcolor{win}52.3 & \cellcolor{lose}33.9 & \cellcolor{lose}33.6 & \cellcolor{lose}69.4 & \cellcolor{win}50.0 & \cellcolor{lose}56.2 & \cellcolor{win}43.3 & \cellcolor{win}51.8 & \textcolor{deltaGreen}{+0.0} \\ % AVG is 51.8, same as baseline
LoRA (r=32)    & \cellcolor{win}33.4 & \cellcolor{win}63.9 & \cellcolor{lose}60.6 & \cellcolor{lose}73.0 & \cellcolor{lose}52.1 & \cellcolor{win}39.1 & \cellcolor{lose}34.4 & \cellcolor{lose}69.7 & \cellcolor{lose}49.2 & \cellcolor{lose}55.6 & \cellcolor{lose}36.5 & \cellcolor{lose}51.6 & \textcolor{deltaRed}{-0.2} \\
PiSSA          & \cellcolor{win}34.6 & \cellcolor{win}66.7 & \cellcolor{lose}59.6 & \cellcolor{lose}73.0 & \cellcolor{lose}51.7 & \cellcolor{lose}33.3 & \cellcolor{lose}33.4 & \cellcolor{lose}69.4 & \cellcolor{win}50.2 & \cellcolor{win}56.3 & \cellcolor{lose}36.5 & \cellcolor{lose}51.3 & \textcolor{deltaRed}{-0.5} \\
\thinmidrule
\\[-0.8em]
FT             & \cellcolor{win}35.5 & \cellcolor{win}65.6 & \cellcolor{lose}60.4 & \cellcolor{lose}74.0 & \cellcolor{win}52.3 & \cellcolor{lose}37.4 & \cellcolor{lose}34.0 & \cellcolor{win}70.1 & \cellcolor{win}50.8 & \cellcolor{win}56.7 & \cellcolor{lose}36.5 & \cellcolor{win}52.1 & \textcolor{deltaGreen}{+0.3} \\
GaLore         & \cellcolor{win}35.2 & \cellcolor{win}65.3 & \cellcolor{lose}58.2 & \cellcolor{lose}74.0 & \cellcolor{lose}52.2 & \cellcolor{lose}34.4 & \cellcolor{lose}33.6 & \cellcolor{win}70.2 & \cellcolor{win}49.7 & \cellcolor{win}56.4 & \cellcolor{lose}36.5 & \cellcolor{lose}51.4 & \textcolor{deltaRed}{-0.4} \\ % AVG 51.4 < 51.8 Baseline
APOLLO         & \cellcolor{win}35.3 & \cellcolor{win}65.7 & \cellcolor{lose}58.0 & \cellcolor{lose}72.0 & \cellcolor{win}52.3 & \cellcolor{lose}34.7 & \cellcolor{lose}34.0 & \cellcolor{win}70.2 & \cellcolor{win}50.3 & \cellcolor{win}57.0 & \cellcolor{lose}36.5 & \cellcolor{lose}51.5 & \textcolor{deltaRed}{-0.3} \\ % AVG 51.5 < 51.8 Baseline
\\[-0.8em]
\thinmidrule
\\[-0.8em]
Ours           & \cellcolor{win}34.8 & \cellcolor{win}65.2 & \cellcolor{lose}56.5 & \cellcolor{lose}73.0 & \cellcolor{win}52.3 & \cellcolor{win}40.1 & \cellcolor{win}35.4 & \cellcolor{win}70.1 & \cellcolor{win}51.6 & \cellcolor{win}57.6 & \cellcolor{win}61.5 & \cellcolor{win}\textbf{54.4} & \textcolor{deltaGreen}{\textbf{+2.6}} \\
\regmidrule
% --- Llama-3.1-8B ---
\\[-0.8em]
\textbf{\textit{Llama-8B}}       & 53.6 & 81.1 & 82.1 & 87.0 & 79.0 & 49.7 & 45.0 & 81.3 & 51.9 & 73.3 & 59.6 & 67.6 & - \\
\thinmidrule
\\[-0.8em]
LoRA (r=8)     & \cellcolor{lose}53.1 & \cellcolor{lose}79.5 & \cellcolor{lose}78.3 & \cellcolor{win}89.0 & \cellcolor{win}80.4 & \cellcolor{win}62.1 & \cellcolor{lose}44.8 & \cellcolor{lose}80.5 & \cellcolor{win}57.8 & \cellcolor{win}74.9 & \cellcolor{lose}54.8 & \cellcolor{win}68.7 & \textcolor{deltaGreen}{+1.1} \\
FT     & \cellcolor{lose}52.3 & \cellcolor{lose}80.2 & \cellcolor{lose}80.5 & \cellcolor{win}91.0 & \cellcolor{win}80.6 & \cellcolor{win}60.8 & \cellcolor{win}45.6 & \cellcolor{lose}81.1 & \cellcolor{win}58.8 & \cellcolor{win}73.7 & \cellcolor{lose}57.7 & \cellcolor{win}69.3 & \textcolor{deltaGreen}{+1.7} \\
\\[-0.8em]
\thinmidrule
\\[-0.8em]
Ours         & \cellcolor{lose}52.1 & \cellcolor{lose}79.5 & \cellcolor{lose}79.6 & \cellcolor{win}91.0 & \cellcolor{win}80.4 & \cellcolor{win}58.6 & \cellcolor{win}46.4 & \cellcolor{lose}80.5 & \cellcolor{win}58.3 & \cellcolor{win}75.2 & \cellcolor{win}64.4 & \cellcolor{win}\textbf{69.6} & \textcolor{deltaGreen}{\textbf{+2.0}} \\
\regmidrule
\end{tabular}
}
\vspace{-1.5em}
\label{tab:main}
\end{table}

% \paragraph{Setup} We apply \dex~to multiple models of varying sizes: Llama-3.1-8B~\cite{grattafiori2024llama}, Llama-3.2-3B/1B~\cite{meta_llama3_2_announcement}, and Qwen-2.5-1.5B/0.5B~\cite{yang2024qwen2}. Since the original pretraining data for these models is not publicly available, we construct a custom corpus comprising web pages, academic papers, encyclopedia entries, and code texts sourced from open datasets~\cite{li2024datacomplm,weifinetuned} roughly following the recipe of OLMo~\cite{olmo20242}. This corpus contains approximately 887 million tokens (using the Llama-3 tokenizer), representing less than 0.01\% of the original pretraining data size for these models. 
% Although \dex~itself is not presented as a fine-tuning method, we compare against baselines trained on the \textit{same} data, including parameter-efficient tuning (LoRA~\cite{hu2022lora}, PiSSA~\cite{meng2024pissa}) and full fine-tuning (FT, using Galore~\cite{zhao2024galore} and APOLLO~\cite{zhu2024apollo})
% to control for the influence of this corpus.
% Comparison with the original \diff~is inevitably incomplete due to absence of pretrained weights.
% We defer full details to Appendix.

\paragraph{Setup} We apply \dex~to Llama-3.1-8B~\cite{grattafiori2024llama}, Llama-3.2-3B/1B~\cite{meta_llama3_2_announcement}, and Qwen-2.5-1.5B/0.5B~\cite{yang2024qwen2}. As the original pretraining data for these models is unavailable, we build a custom corpus of web pages, papers, encyclopedias, and code from open datasets~\cite{li2024datacomplm,weifinetuned}, similar to OLMo~\cite{olmo20242}. This corpus contains 887M tokens (Llama-3 tokenizer), less than 0.01\% of the models' original pretraining data size.
Although \dex~is not presented as a fine-tuning method, we compare against baselines trained on the \textit{same} data—including parameter-efficient tuning (PEFT; LoRA~\cite{hu2022lora}, PiSSA~\cite{meng2024pissa}) and full fine-tuning (FT; Galore~\cite{zhao2024galore}, APOLLO~\cite{zhu2024apollo})—to control for the influence of this corpus.
Direct comparison with original \diff~is limited by unavailable pretrained weights, and we defer evaluations in smaller settings to Appendix along with other details. 
For head selection we simply set $k$ to be half the total number of heads for each model, and we adopt the $\lambda_{init}$ from \cite{ye2024differential} (we provide ablations in Appendix~\ref{supp:2.3}). For PiSSA we report $r=32$ case as this yields good results.
% For LoRA we report results for two different ranks (8 and 32), and for PiSSA, we report $r=32$ case as this yields good overall performance.
% Full details are in the Appendix.

\vspace{-0.5em}

\paragraph{Results} We report performances on 11 widely used language modeling benchmarks~\cite{Clark2018ThinkYH,NEURIPS2019_4496bf24,wang-etal-2018-glue,zellers2019hellaswag,OpenBookQA2018,Bisk2020,sakaguchi2019winogrande} using \cite{eval-harness}. As shown in Table~\ref{tab:main}, \dex~achieves significant improvements across model sizes and families. Given the discrepancy between our training corpus, original pretraining data and the downstream tasks, it is natural to observe degradation after additional training in some cases. 
% Nevertheless, \dex~demonstrates robust performance gains on the majority of benchmarks, even when other methods exhibit performance drops. 
Nevertheless, \dex~demonstrates robust performance gains on the majority of benchmarks, even when other methods—all trained on the same corpus—exhibit performance drops.
In particular, we attribute \dex's strong performance on WSC to its enhanced anaphora resolution granted by the capacity to model negative relevance for incorrect antecedents, which aligns with our intuitions.
Notably, although \dex~only updates self-attentions, it consistently outperforms both PEFT and full fine-tuning even when full tuning steadily outperforms PEFT (\textit{e.g.,} Llama-1B). 

% These results underscore the effectiveness of \dex~in transferring the general strengths of \diff~to pretrained LLMs.

\subsection{Key Information Retrieval}
\label{sec:4.2}

Needle-in-a-Haystack test~\cite{LLMTest_NeedleInAHaystack} is widely adopted to assess LLM's ability to identify critical information embedded in an extensive context. Following the multi-needle retrieval setting of \cite{liu2024world,team2024gemini,ye2024differential}, we 
% insert target "needles"—sentences assigning unique numbers to specific cities—at varying depths within contexts of different lengths. The retrieval task requires the models to identify the correct number corresponding to queried city. We 
place the needle at five distinct depths within the context: 0\%, 25\%, 50\%, 75\%, and 100\%, accompanied by distracting needles. We note the total number of needles placed in the context as $N$, and the number of target needles actually being queried as $R$.
Each combination of depth and context length is assessed using 20 samples. 

\begin{figure}[t]
    \centering
    \begin{subfigure}[b]{0.475\textwidth}
        \centering
        \includegraphics[width=\textwidth]{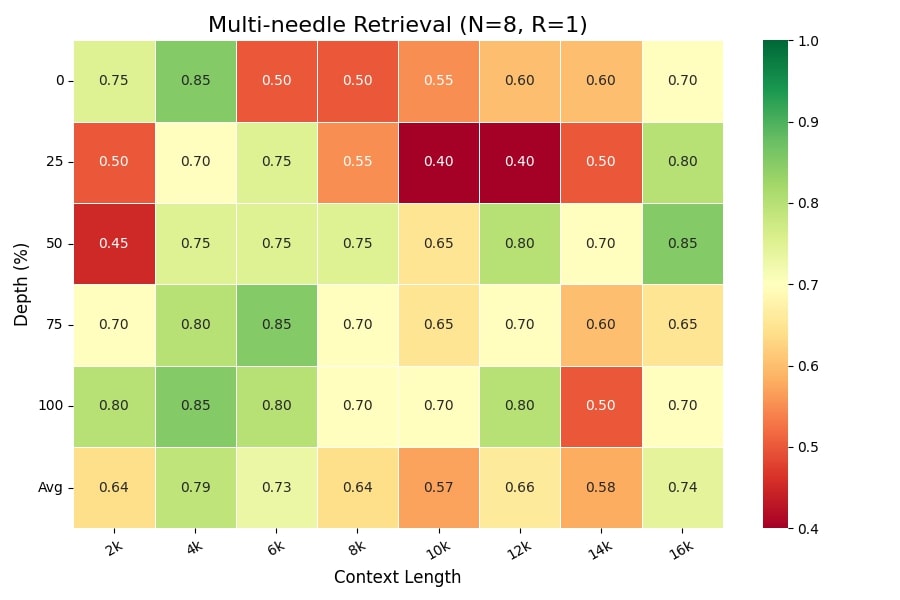}
        \captionsetup{font=footnotesize} % or scriptsize, footnotesize, etc.
        % \vspace{-1em}
        \caption{LLAMA}
        \label{fig:head_attn:a}
    \end{subfigure}
    \hspace{0.5em}
    \begin{subfigure}[b]{0.475\textwidth}
        \centering
        \includegraphics[width=\textwidth]{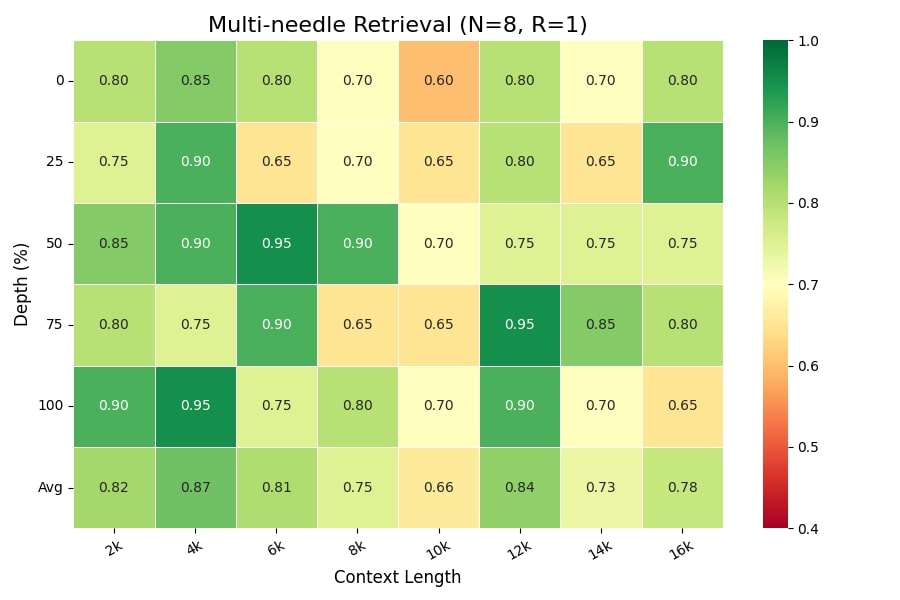}
        \captionsetup{font=footnotesize} % or scriptsize, footnotesize, etc.
        % \vspace{-1em}
        \caption{LLAMA + \dex}
        \label{fig:head_attn:b}
    \end{subfigure}
    \caption{Multi-needle retrieval results. $N$: total number of needles, $R$: number of queries.}
    \label{fig:needle}
    \vspace{-1em}
\end{figure}

Fig.\ref{fig:needle} shows the result for $N=8, R=1$ case. \dex~significantly enhances the retrieval performance of the base Llama-3B model across all context lengths and embedding depths. Notably, \dex~improves the average accuracy score by 11.4\% (absolute increase from 66.9\% to 78.3\%), highlighting its effectiveness in improving key information retrieval capabilities.

\begin{wraptable}{r}{0.45\textwidth} 
    \centering 
    \vspace{-1.1em} 
    \caption{Effective attention scores allocated to the answer spans inserted at different depths in key information retrieval.}
    \label{tab:noise_cancel} 
    \resizebox{0.45\textwidth}{!}{
        \begin{tabular}{l c c c c c| c} 
            \specialrule{1.5pt}{0pt}{0.3em} 
            % \vspace{0.5em}
            \multicolumn{1}{c}{} & \multicolumn{6}{c}{\textbf{Attention to Answer $\uparrow$}} \\
            \cmidrule(lr){2-7}
            
            Model & 0\% & 25\% & 50\% & 75\% & 100\% & Avg \\
            \midrule 
            Llama & 0.06 & 0.04 & 0.06 & 0.05 & 0.08 & 0.06 \\
            \dex   & \textbf{0.21} & \textbf{0.13} & \textbf{0.18} & \textbf{0.16} & \textbf{0.27} & \textbf{0.19} \\
            \bottomrule
        \end{tabular}}
    \vspace{-1em} 
\end{wraptable}

Increased attention to answer ratio in Table \ref{tab:noise_cancel} further demonstrates that \dex~effectively transfers the core capability of \abbr~attention: attention noise canceling.
Despite being implicitly applied to the output value matrix, \dex~notably alters the \textit{effective attention pattern} to focus on relevant information. This result empirically supports our design choice, placing \dex~on the sweetspot between efficiency and efficacy.
We provide details in Appendix \ref{supp:5.3}.

\subsection{In-Context Learning}
\label{sec:4.3}

\begin{wraptable}{r}{0.45\textwidth}
\vspace{-1.7em}
\centering
\caption{In-context learning performance.}
\setlength\intextsep{0pt}
\resizebox{0.45\textwidth}{!}{%
\setlength{\tabcolsep}{1.2mm}{
\begin{tabular}{l rrrrrr r}
    \specialrule{1.5pt}{0pt}{0.3em} 
    & \multicolumn{6}{c}{N-shot} & \\
    \cmidrule(lr){2-7}
    Dataset & 1 & 10 & 100 & 500 & 1000 & 2000 & Avg \\ 
    \midrule
    \multicolumn{8}{l}{\textit{TREC}} \\
    \quad Llama & 20.0 & 71.1 & 88.9 & 93.3 & 88.9 & 93.3 & 75.9 \\
    \quad LoRA & 20.0 & 68.9 & 93.3 & 93.3 & 91.1 & 93.3 & 76.7 \\
    \quad FT & 16.0 & 76.0 & 86.0 & 92.4 & 91.1 & 93.3 & 75.8 \\
    \quad \dex & 26.7 & 84.4 & 86.7 & 93.3 & 88.9 & 93.3 & \textbf{78.9} \\
    \midrule
    \multicolumn{8}{l}{\textit{Banking-77}} \\
    \quad Llama & 24.4 & 35.6 & 55.6 & 86.7 & 91.1 & 91.1 & 64.1 \\
    \quad LoRA & 26.7 & 40.0 & 53.3 & 88.9 & 88.9 & 91.1 & 64.8 \\
    \quad FT & 21.6 & 34.4 & 56.0 & 84.4 & 88.8 & 92.4 & 62.9 \\
    \quad \dex & 22.2 & 37.8 & 60.0 & 91.1 & 95.6 & 95.6 & \textbf{67.0} \\
    \midrule
    \multicolumn{8}{l}{\textit{Clinic-150}} \\
    \quad Llama & 15.6 & 44.4 & 60.0 & 82.2 & 95.6 & 95.6 & 65.6 \\
    \quad LoRA & 22.2 & 42.2 & 57.8 & 82.2 & 95.6 & 95.6 & 65.9 \\
    \quad FT & 22.2 & 42.2 & 60.0 & 82.2 & 93.3 & 95.6 & 65.9 \\
    \quad \dex & 22.2 & 40.0 & 57.8 & 82.2 & 97.8 & 97.8 & \textbf{66.3} \\
    \bottomrule
\end{tabular}}}
\vspace{-3.2em}
\label{tab:icl_results}
\end{wraptable}

\diff~notably enhances in-context learning performance compared to standard transformer models. To validate whether \dex~can achieve similar improvements, we conduct a comprehensive evaluation using three established benchmarks: TREC~\cite{hovy2001toward}, Banking-77~\cite{casanueva2020efficient}, and Clinic-150~\cite{larson2019evaluation}. Following the setup of \cite{bertsch2024context}, we adopt a random selection procedure for N-shot examples, as retrieval-based scores quickly saturate with state-of-the-art LLMs.

The results summarized in Table~\ref{tab:icl_results} clearly illustrate that \dex~consistently delivers performance gains across all evaluated benchmarks compared to both the base Llama model and fine-tuning baselines (LoRA and FT). \dex~achieves the highest average accuracy across varying N-shot settings, demonstrating its robustness and efficacy in enhancing the in-context learning capabilities of pretrained models.

\subsection{Application to Instruction Tuning}
\label{supp:3}

We investigate whether \dex~can likewise enhance performance on instruction-following tasks. 
% Consequently, the experimental protocol established for base pretrained models cannot be directly transferred to the instruction-following scenario without modification.
To fairly assess the effect of \dex, we adopt two complementary settings.
First, we apply \dex~on a publicly available instruction-tuned checkpoint trained on an open-source instruction corpus OpenHermes-2.5~\cite{openhermes25}\footnote{\url{https://huggingface.co/artificialguybr/Meta-Llama-3.1-8B-openhermes-2.5}} using the same training data (OH-2.5). This \textit{continued} instruction tuning setting eliminates the confounding effect of training data and lets us verify whether \dex~improves the performance of an existing instruct model.
Second, we directly apply \dex~to a base pretrained model as an instruction-tuning method itself, similarly using OH-2.5 but in single stage.
We examine if \dex~can effectively induce instruction-following capabilities without prior end-to-end instruction tuning. Note that we include FT (further fine-tuning the open-source checkpoint on the same OpenHermes data for more steps) as an additional baseline to alleviate the concern for underfitting, which clearly distinguishes the contribution of \dex~from the benefit of more training steps.

\begin{table}[t]
\centering
\caption{\textbf{Instruction-tuning results on 8 benchmarks.} 
The top four rows correspond to the first setting, while the bottom two rows correspond to the second.
% All values are rounded to one decimal place. 
}
\vspace{0.3em}
\label{tab:instruct}
\resizebox{0.9\textwidth}{!}{%
\begin{tabular}{lcccccccc|cc}
\toprule
\textbf{Model} & \textbf{MMLU} & \textbf{Arc-C} & \textbf{IFEval} & \textbf{MBPP++} & \textbf{GSM8K} & \textbf{AGIEval} & \textbf{HumanEval} & \textbf{Math500} & \textbf{AVG} & \textbf{$\Delta$} \\
\regmidrule
\\[-0.8em]
\multicolumn{7}{l}{\textbf{\textit{Instruction-tuned}}} \\
\\[-0.8em]
Base       & 62.9 & 78.3 & 46.8 & 68.3 & 71.1 & 32.2 & 44.5 & 13.4 & 52.2 & - \\
\\[-0.8em]
+ LoRA     & \cellcolor{win}63.1 & \cellcolor{win}79.5 & \cellcolor{lose}45.7 & \cellcolor{lose}65.3 & \cellcolor{lose}70.3 & \cellcolor{win}40.6 & \cellcolor{win}47.0 & \cellcolor{lose}4.0 & \cellcolor{lose}51.9 & \textcolor{deltaRed}{-0.3} \\
\\[-0.8em]
+ FT & \cellcolor{win}63.0 & \cellcolor{win}78.6 & \cellcolor{win}49.2 & \cellcolor{lose}63.2 & \cellcolor{lose}68.8 & \cellcolor{win}42.3 & \cellcolor{lose}36.6 & \cellcolor{win}20.0 & \cellcolor{win}53.7 & \textcolor{deltaGreen}{+1.5} \\
\\[-0.8em]
+ \dex   & \cellcolor{win}63.1  & \cellcolor{lose}77.7  & \cellcolor{win}57.2  & \cellcolor{lose}64.8  & \cellcolor{win}74.3  & \cellcolor{win}40.7  & \cellcolor{win}47.6 & \cellcolor{win}19.2 & \cellcolor{win}55.6  & \textcolor{deltaGreen}{\textbf{+3.4}} \\
\regmidrule
\multicolumn{9}{l}{\textbf{\textit{Pretrained}}} \\
\\[-0.8em]
+ LoRA & \cellcolor{win}63.7 & \cellcolor{lose}70.5 & \cellcolor{lose}42.0 & \cellcolor{lose}65.3 & \cellcolor{lose}57.4 & \cellcolor{win}35.4 & \cellcolor{win}45.7 & \cellcolor{lose}2.0 & \cellcolor{lose}47.8 & \textcolor{deltaRed}{-4.4}\\
\\[-0.8em]
+ \dex & \cellcolor{win}63.6 & \cellcolor{lose}77.3 & \cellcolor{win}51.0 & \cellcolor{lose}66.1 & \cellcolor{lose}68.4 & \cellcolor{win}37.9 & \cellcolor{win}50.7 & \cellcolor{win}16.2 & \cellcolor{win}53.9 & \textcolor{deltaGreen}{\textbf{+1.7}} \\
\regmidrule
\end{tabular}
}
\end{table}

Table~\ref{tab:instruct} reports results on eight representative benchmarks that span language understanding~\cite{hendrycks2020measuring}, commonsense reasoning~\cite{Clark2018ThinkYH}, instruction following~\cite{zhou2023instruction}, math~\cite{cobbe2021training,hendrycks2020measuring}, code generation~\cite{austin2021program,chen2021evaluating}, and general human task~\cite{zhong2023agieval}. We observe that \dex~delivers favorable results on diverse settings, significantly outperforming baselines on benchmarks like GSM8K, HumanEval and IFEval.
When directly applied to a base pretrained model, \dex~achieves notably higher performance than LoRA, demonstrating comparable performance to more heavily tuned baselines (top 3 rows) without any end-to-end SFT.
These results indicate \dex's effectiveness in inducing and reinforcing instruction-following capabilities efficiently.

\subsection{Ablation and Analysis}
\label{sec:4.4}

We conduct ablation experiments using Llama-3B model. We mainly focus on two critical components: head selection strategies and learnable lambda annealing. We report the average score for the 11 language modeling benchmarks (similar to Table \ref{tab:main}). Appendix \ref{supp:2} presents full results.

\begin{wraptable}{r}{0.49\textwidth}
% \vspace*{2\baselineskip}
\centering
\vspace{-1.2em}
\caption{Ablation with head selection and lambda control strategies. \textbf{imp.} refers to importance-based and \textbf{ent.} stands for entropy-based.}
\resizebox{0.49\textwidth}{!}{%
\begin{tabular}{l| c c c| c}
    \specialrule{1.5pt}{0pt}{0pt}
    \\[-0.8em]
    \textbf{Model} & \textbf{Head Selection} & \textbf{$\lambda$-learned} & \textbf{$\lambda$-annealed} & \textbf{LM Acc (\%)} \\
    \midrule
    Llama & - & - & - & 60.2 \\
    \dex & all & \ding{51} & \ding{51} & 61.9 \\
    \dex & imp. & \ding{51} & \ding{51} & 63.9 \\
    \dex & ent. {\scriptsize($\downarrow$)} & \ding{51} & \ding{51} & 62.8 \\
    % \rowcolor{lose}
    \rowcolor{gray!20}
    \dex & ent. {\scriptsize($\uparrow$)} & \ding{51} & \ding{51} & \textbf{64.2} \\
    \midrule
    \dex & ent. & \ding{51} & \ding{55} & 63.8 \\
    \dex & ent. & \ding{55} & \ding{51} & 63.4 \\
    \dex & ent. & \ding{55} & \ding{55} & 62.4 \\
    \specialrule{1pt}{0pt}{0pt}
\end{tabular}}
\vspace{-1.2em}
\label{tab:head_lambda_results}
\end{wraptable}

From Table~\ref{tab:head_lambda_results}, it is evident that incorporating entropy-based head selection combined with both learnable and annealed lambda methods yields the best performance, achieving the overall accuracy 
of 64.2\%. Removing either component from lambda leads to noticeable performance drops, indicating the necessity of both. Additionally, both head selection strategies outperform the configuration without head selection, with the entropy-based strategy pushing the boundary further. The fact that choosing low entropy heads ($\downarrow$) underperforms further supports our design.
These findings underline the complementary roles of the head selection and lambda annealing mechanisms in maximizing the effectiveness of \dex.

% ($||\lambda f_D (\mathbf{O})|| / ||\mathbf{O}||$)
% First we study how \dex~modifies the internal representation by looking at the cosine similarity and relative norm of $\mathbf{O}$ and $\lambda f_D (\mathbf{O})$. 
% (2) Different heads exhibit distinct modulation strategies; the head on the left applies its strongest changes during amplification (negative cosine), while the head on the right focuses its strongest changes on attenuation (positive cosine), suggesting functional specialization.
% Second, we revisit attention head CKA (Fig.\ref{fig:analysis:b}). Compared to Llama baseline (top row), \dex~shows significantly diminished redundancy between attention heads (bottom row). 

\begin{figure}[h]
    \centering
    \begin{subfigure}[b]{0.48\linewidth}
        \centering
        \includegraphics[width=\textwidth]{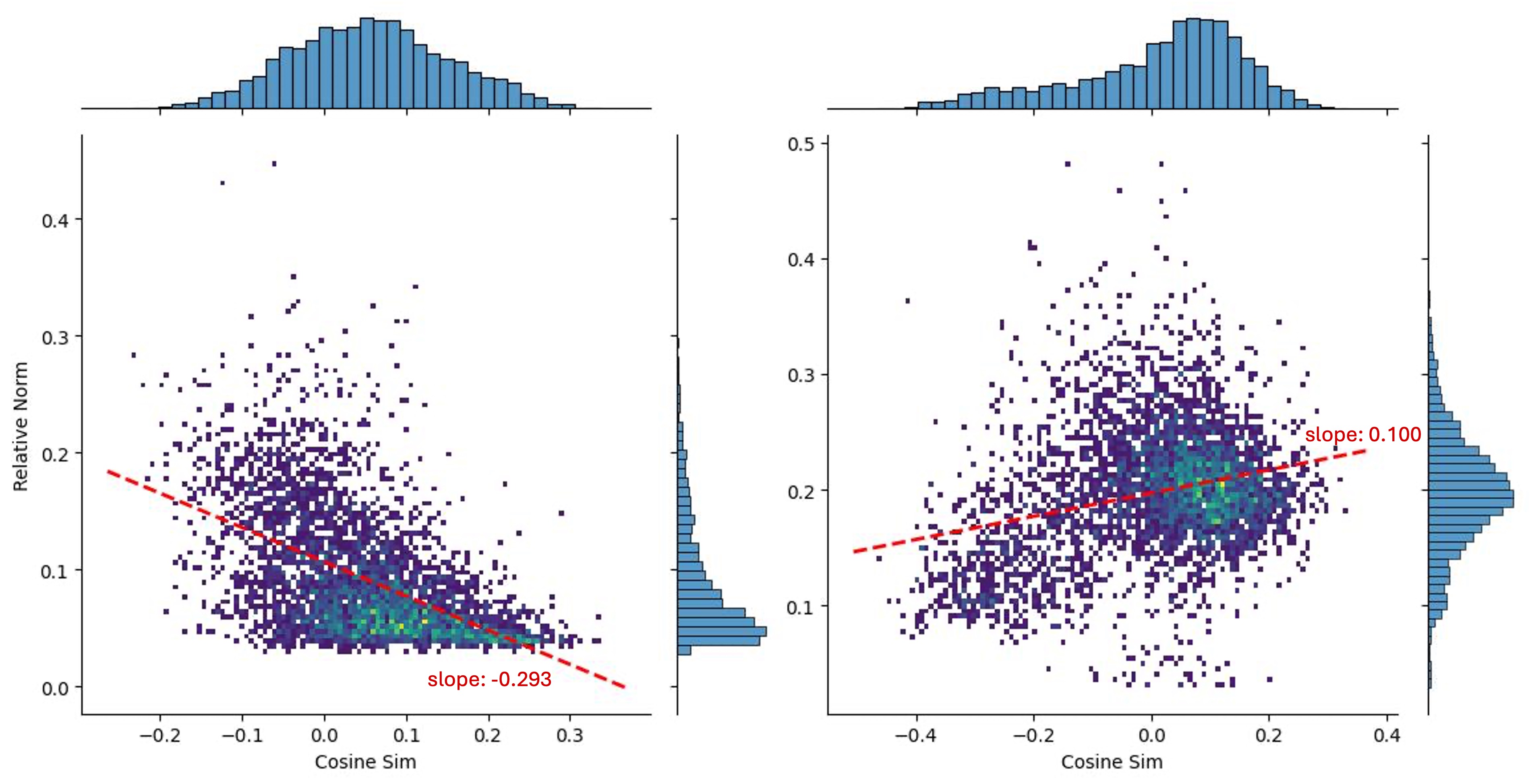}
        \captionsetup{font=footnotesize} % or scriptsize, footnotesize, etc.
        \caption{Jointplot between $\mathbf{O}$ and $\lambda f_D (\mathbf{O})$.}
        \label{fig:analysis:a}
    \end{subfigure}
    \hfill
    \begin{subfigure}[b]{0.21\linewidth}
        \centering
        \includegraphics[width=\textwidth]{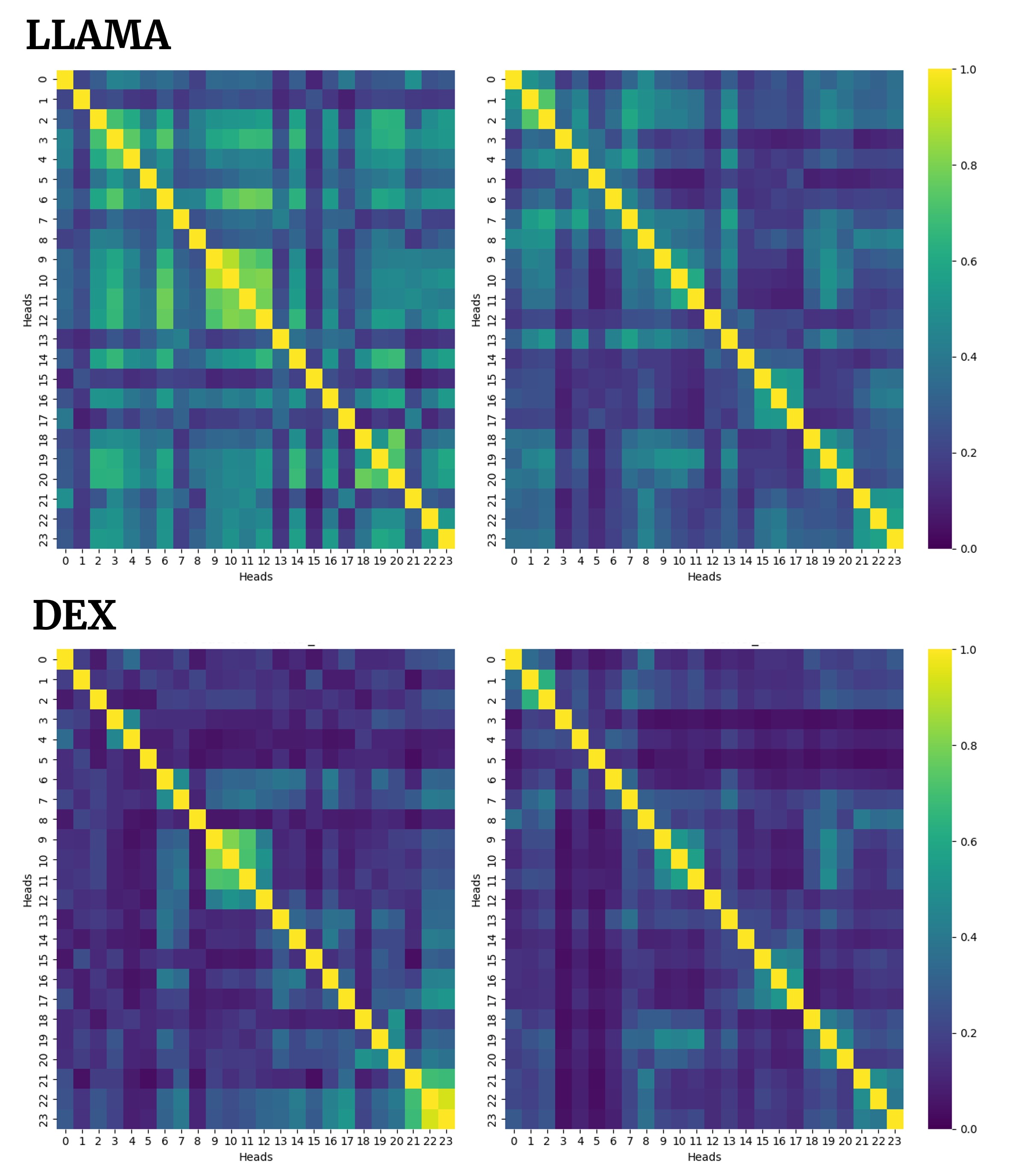}
        \captionsetup{font=footnotesize} % or scriptsize, footnotesize, etc.
        \caption{Head CKA.}
        \label{fig:analysis:b}
    \end{subfigure}
    \hfill
    \begin{subfigure}[b]{0.23\linewidth}
        \centering
        \includegraphics[width=\textwidth]{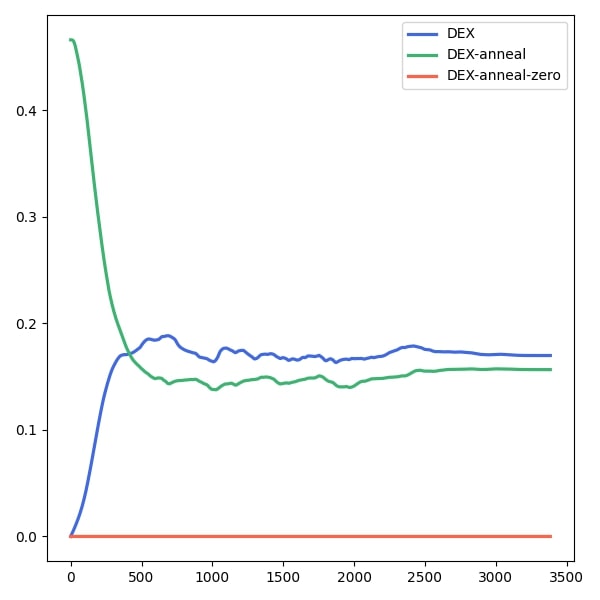}
        \captionsetup{font=footnotesize} % or scriptsize, footnotesize, etc.
        \caption{$\lambda$ during training.}
        \label{fig:analysis:c}
    \end{subfigure}
    \caption{\textbf{Analysis on \dex}. 
    % (a) shows the cosine similarity and relative norm from two different heads of the same layer. 
    (a) Cosine similarity ($\mathrm{cosine}(\mathbf{O}, \Delta)$) vs. Relative Norm ($||\Delta|| / ||\mathbf{O}||$, where $\Delta = \lambda f_D (\mathbf{O})$)
    (b) CKA of attention head output features ({\color{viridislight}brighter} means higher redundancy). 
    (c) Training dynamics of learnable $\lambda$ under different initialization/annealing schemes.
    }
    \label{fig:analysis}
    \vspace{-0.5em}
\end{figure}

We also analyze the inner working mechanism of \dex. 
First, we investigate how \dex~modifies the original attention output $\mathbf{O}$ via the subtracted term $\Delta = \lambda f_D (\mathbf{O})$. 
% We plot the cosine similarity against the relative norm (Fig.\ref{fig:analysis:a}).
% Key observations include:
% (1) The cosine similarity is broadly distributed, centered near zero, indicating \dex's capacity to both reinforce and suppress certain features. 
% (2) Heads exhibit distinct strategies, with some focusing maximal changes on amplification (negative cosine, left) and others on attenuation (positive cosine, right), suggesting functional specialization.
% We plot the cosine similarity against relative norm in Fig.\ref{fig:analysis:a}, where the former determines the direction of modification (\textit{e.g.,} positive cosine for suppression) and the latter represents the magnitude. 
% We observe that while the similarity distribution suggests \dex's capacity to both reinforce and suppress features, heads exhibit distinct patterns, with some focusing on amplification (shown by larger norm in negative cosine region, left) and others on attenuation (larger norm for positive cosine, right).
Fig.~\ref{fig:analysis:a} plots cosine similarity (indicating modification direction, e.g., positive for suppression) against relative norm (modification magnitude). 
We observe that while the similarity distribution suggests \dex's capacity to both reinforce and suppress features, heads exhibit distinct patterns, with some focusing on amplification (higher norm for negative cosine, left) and others on attenuation (higher norm for positive cosine, right).
% This reveals \dex's capacity to both reinforce and suppress features, with heads adopting distinct amplification (high norm, negative cosine) or attenuation (high norm, positive cosine) strategies, suggesting emergent functional specialization.
Second, CKA on head output features reveals that \dex~notably reduces inter-head redundancy (Fig.\ref{fig:analysis:b}), implying more diverse head specialization.
Lastly, we monitor $\lambda$ during training in Fig.\ref{fig:analysis:c}. Simply zero-initializing the learnable $\lambda$ ({\color{tomato}red}) completely fails to introduce \dex, while removing annealing ({\color{mediumseagreen}green}) results in instability at the initial phase. Our approach ({\color{royalblue}blue}) smoothly introduces \dex~with minimal damage to the pretrained knowledge. Refer to Appendix for full results.

% We plot the cosine similarity against relative norm in Fig.\ref{fig:analysis:a}, where the former determines the direction of modification (\textit{e.g.,} positive cosine for suppression) and the latter represents the magnitude. 
% We first observe that the cosine similarity is broadly distributed and centered near zero, suggesting \dex's capacity to both reinforce and suppress features. In addition, heads exhibit different patterns, with some focusing on amplification (shown by larger norm in negative cosine region, left) and others on attenuation (larger norm for positive cosine, right).

% \begin{wrapfigure}{r}{0.4\textwidth}
% \vspace{-0.8em}
% \centering
% \setlength\intextsep{0pt}
% \resizebox{0.37\textwidth}{!}{%
%     \includegraphics[width=\textwidth]{figs/efficiency/benchmark_long.jpg}
% }
% \vspace{-0.8em}
% \caption{Throughput comparison.}
% \vspace{-1.5em}
% \label{fig:throughput}
% \end{wrapfigure}

\begin{figure}[h]
    \centering
    \begin{subfigure}[b]{0.48\linewidth}
        \centering
        \includegraphics[width=\textwidth]{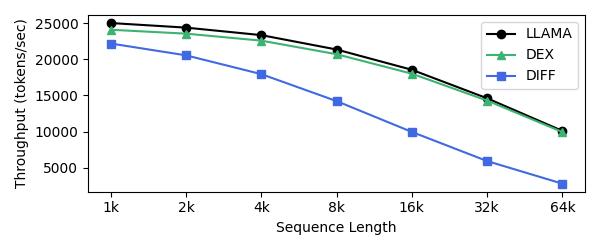}
        \captionsetup{font=footnotesize} % or scriptsize, footnotesize, etc.
        \caption{Throughput comparison.}
        \label{fig:throughput}
    \end{subfigure}
    \hfill
    \begin{subfigure}[b]{0.48\linewidth}
        \centering
        \includegraphics[width=\textwidth]{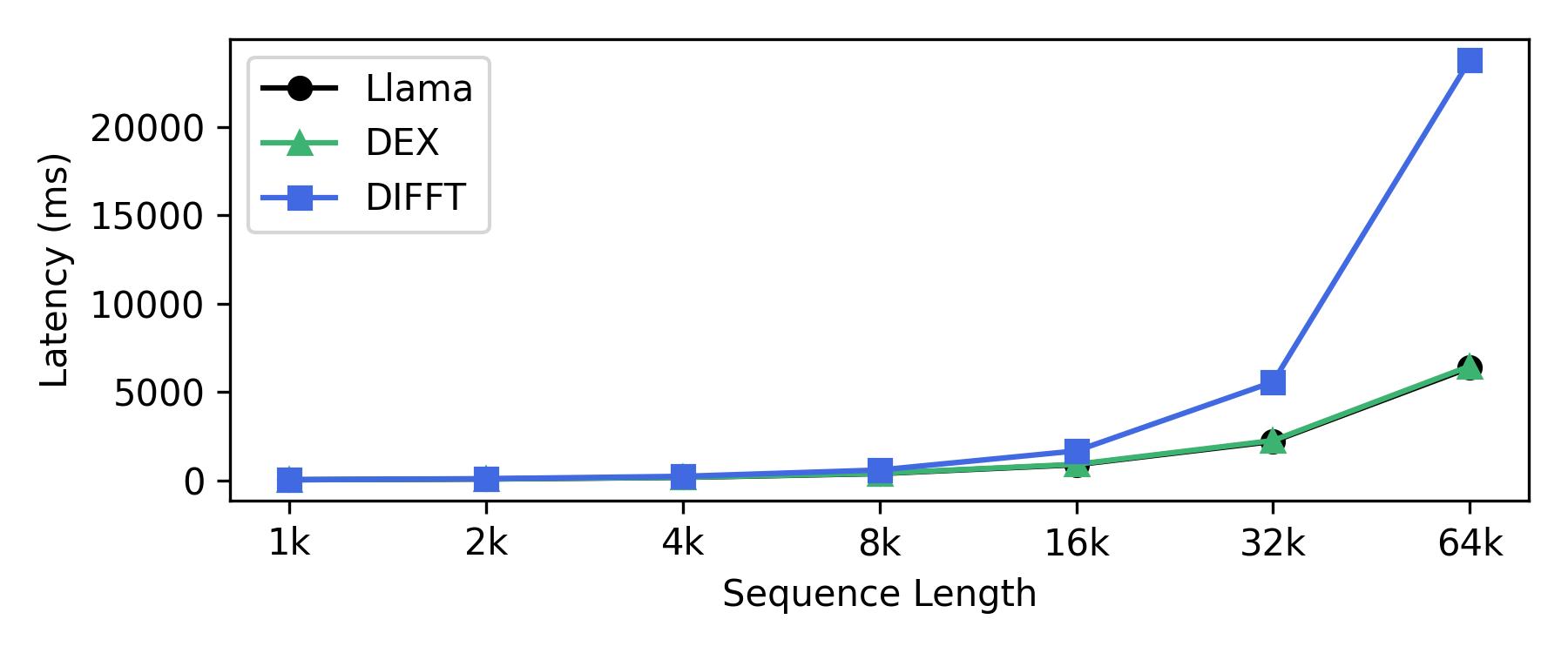}
        \captionsetup{font=footnotesize} % or scriptsize, footnotesize, etc.
        \caption{Latency Comparison.}
        \label{fig:latency}
    \end{subfigure}
    \caption{\textbf{Inference time efficiency analysis.}
    We benchmark (a) throughput and (b) latency of three attention variants. While \abbr~attention costs significantly more compute at test-time compared to the original Llama attention, \dex~incurs negligible overhead thanks to its lightweight design. All benchmarks are measured on a single Nvidia A100 GPU.
    }
    \label{fig:efficiency}
    \vspace{-0.5em}
\end{figure}

We verify the test-time efficiency of \dex~by comparing the throughput (tokens per second) and latency with base Llama and \diff~(3B). Fig.\ref{fig:efficiency} clearly shows that both in terms of throughput and latency, \dex~demonstrates superior inference time efficiency.
While \diff~exhibits increasing inefficiency with longer context due to its compute-heavy attention operation, \dex~remains competitive to the original Llama baseline thanks to its lightweight design.

\begin{wrapfigure}{r}{0.4\textwidth}
\vspace{-1.2em}
\centering
\setlength\intextsep{0pt}
\resizebox{0.37\textwidth}{!}{%
    \includegraphics[width=\textwidth]{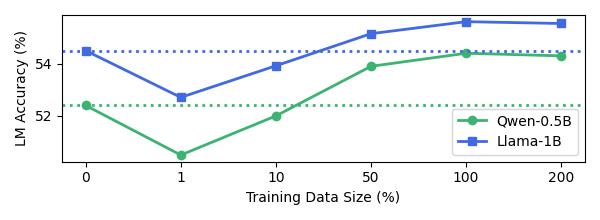}
}
\vspace{-0.2em}
\caption{\dex~with varying data size.}
\vspace{-1.5em}
\label{fig:data_abl}
\end{wrapfigure}

% Finally, we present the performance of \dex~with different training data size. Fig.\ref{fig:data_abl} shows that using as few as 400M tokens delivers non-trivial gains, and the performance plateaus afterwards. 
% This is because our dataset is not directly correlated to the downstream benchmarks. 
% Hence, while we do need certain number of training steps to successfully integrate \dex, simply pouring in more volume of general data does not lead to improved performance.

Finally, we evaluate the effect of training data size on the application of \dex~(Fig. \ref{fig:data_abl}). Notable gains appear with just 400M tokens, highlighting the lightweight nature of \dex. Since our training data lacks direct correlation with the downstream benchmarks, modest amount of data (<1B) is sufficient to elicit the full potential of \dex, and simply adding more general data yields diminishing returns in downstream tasks.
% this suggests that while sufficient data is needed to integrate \dex, simply adding more general data yields diminishing returns for specific tasks.

\section{Conclusion}
\label{sec:5}

In this work, we study the internal mechanism of \diff~to identify three key factors behind its empirical success: enhanced expressivity via negative attention, reduced redundancy among attention heads and improved optimization dynamics. Based on these insights, we propose \dex, an architectural adaptation method that efficiently integrates the empirical strengths of \diff~into pretrained LLMs with standard transformer architecture. Diverse evaluation results confirm the effectiveness and versatility of \dex.
% In this work, we study the internal mechanism of \diff~to better understand what drives its empirical success. Based on these insights, we propose \dex, an effective architectural adaptation method that integrates the strengths of \abbr~attention into pretrained LLMs with lightweight training.
\vspace{-1em}
\paragraph{Limitations} 
% Due to the absence of pretrained weights, our analysis on \diff~uses the model we trained. 
For \diff~analysis,
we followed the original setup as much as possible, but different behaviors can emerge under different model scale, data composition, etc. Similarly, though \dex~works well across model sizes, it has not been tested beyond 8B parameter scale. We leave it for future works.
\vspace{-1em}
% we have not tested its effectiveness beyond 8B parameter scale.
\paragraph{Acknowledgement}
N. Kwak was supported by NRF (2021R1A2C3006659) and IITP grants (RS-2021-II211343, RS-2022-II220320, RS-2025-25442338), all funded by the Korean Government.
% we have not tested its effectiveness beyond 8B parameter scale.

% We hope our findings facilitate future research in LLM architectures, which we believe is still a rather open research question.

%%%%%%%%%%%%%%%%%%%%%%%%%%%%%%%%%%%%%%%%%%%%%%%%%%%%%%%%%%%%

\bibliographystyle{plainnat}  % Specifies the bibliography style, plainnat is a natbib-compatible style
\bibliography{references}     % The filename of the BibTeX file, without the extension
%%%%%%%%%%%%%%%%%%%%%%%%%%%%%%%%%%%%%%%%%%%%%%%%%%%%%%%%%%%%

%%%%%%%%%%%%%%%%%%%%%%%%%%%%%%%%%%%%%%%%%%%%%%%%%%%%%%%%%%%%
% \newpage
% \appendix
% \input{suppl}

%%%%%%%%%%%%%%%%%%%%%%%%%%%%%%%%%%%%%%%%%%%%%%%%%%%%%%%%%%%%

\newpage
\appendix
% \begin{wrapfigure}{r}{0.44\textwidth}
% \vspace{-1.2em}
% \centering
% \setlength\intextsep{0pt}
% \begin{subfigure}[b]{0.2099\textwidth} % Half of 0.42
%     \centering
%     \includegraphics[width=\linewidth]{figs/abs/spearman_abs.png}
%     \vspace{-1.7em}
%     \caption{Rank corr.}
%     \label{fig:abs:a}
% \end{subfigure}%
% \hfill
% \begin{subfigure}[b]{0.2099\textwidth}
%     \centering
%     \includegraphics[width=\linewidth]{figs/abs/pearson_abs.png}
%     \vspace{-1.7em}
%     \caption{Pearson's corr.}
%     \label{fig:abs:b}
% \end{subfigure}
% \vspace{-0.3em}
% \caption{Correlations between softmax and \abbr~attention, with and without abs().}
% \vspace{-1em}
% \label{fig:abs}
% \end{wrapfigure} 

\definecolor{royalblue}{RGB}{65,105,225}
\definecolor{mediumseagreen}{RGB}{60,179,113}
\definecolor{tomato}{RGB}{255,99,71}

\section{Related Works}
\label{supp:1}
\textbf{Differential Transformer}~\cite{ye2024differential} introduces an architecture designed to mitigate attention noise~\cite{LLMTest_NeedleInAHaystack,liu2024lost,lu2021fantastically}, a known challenge in transformer models. Its core mechanism involves computing attention scores using two groups and then subtracting the resulting attention maps, aiming to cancel out common-mode noise components. Building on this, DINT Transformer~\cite{cang2025dinttransformer} aims to improve numerical stability and training dynamics by incorporating an integral term alongside the differential one. However, these pioneering works do not provide detailed mechanistic analyses explaining \textit{why} differential attention is effective. Furthermore, both architectures inherently require computing two separate attention pathways, resulting in substantial computational overhead compared to standard attention. This increased cost hinders practical deployment, particularly for large-scale models. Building on top of our analysis on \diff's success, we propose \dex~that implicitly embeds the benefits of differential attention into pretrained language models without nontrivial computational overhead.

\textbf{Negative Attention Scores.} Several approaches explicitly introduce negative attention scores. Centered Attention~\cite{ali2023centered}, for instance, adds offsets to the softmax calculation, forcing attention weights per query to sum to zero (rather than one) to mitigate over-smoothing. Other methods achieve negative weighting through direct manipulations of the softmax operation or via linear attention approximations~\cite{lv2024more, meng2025polaformer}, often demonstrating enhanced representational expressivity. However, methods that explicitly alter the core attention computation can introduce training stability challenges and often lack compatibility with highly optimized implementations like FlashAttention~\cite{dao2022flashattention}. In contrast, \dex~aims to capture the benefits of signed, differential attention implicitly. By applying its learnable transformation \textit{after} the standard softmax attention calculation (i.e., to the output values), \dex~avoids modifying the core QK-softmax pathway, thereby maintaining compatibility and potentially simplifying integration and training.

\textbf{Attention Redundancy and Entropy.} Numerous works~\cite{cheng2024mini,molchanov2019importance,michel2019sixteen, voita2019analyzing, dingpass,li2021differentiable, xia2023sheared, bian2021attention} have shown that there is significant redundancy among attention heads in multi-head attention, and propose head pruning methods to enhance efficiency. Instead of getting rid of unimportant heads, our approach applies implicit differential adaptation to redundant heads, effectively revitalizing them and modeling richer attention representations. Meanwhile, attention entropy-based analyses have provided insights into the transformer attention mechanisms. \cite{zhang2024attention, meng2025polaformer} argues that excessively high attention entropy negatively impacts performance, while \cite{jha2025entropy, zhai2023stabilizing} associates entropy with training stability. 
In this work, we leverage attention entropy in two ways: (1) understanding the attention score distribution (and potential sparsity), and (2) identifying less critical attention heads.

\section{Ablation and Analysis}
\label{supp:2}

In this section, we present additional empirical results to support our design choice and analysis.

\subsection{Attention Magnitude Correlation}
\label{supp:2.1}

\begin{wrapfigure}{r}{0.44\textwidth}
\vspace{-1.2em}
\centering
\setlength\intextsep{0pt}
\begin{subfigure}[b]{0.2099\textwidth} 
    \centering
    \includegraphics[width=\linewidth]{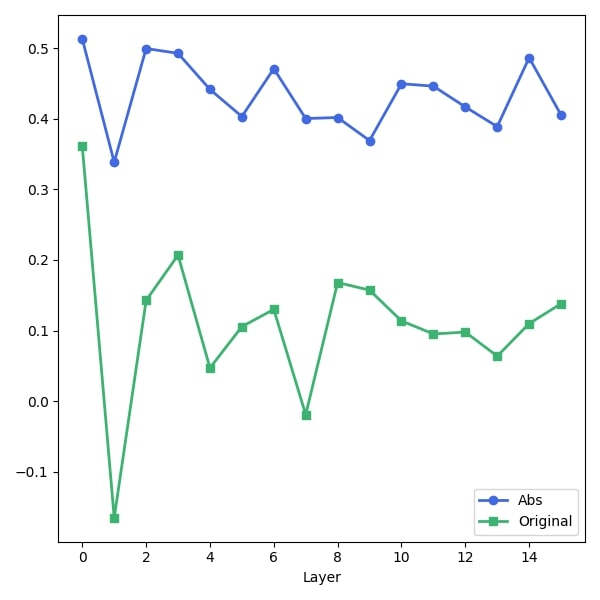}
    \vspace{-1.7em}
    \caption{Rank Corr.}
    \label{fig:magnitude:a}
\end{subfigure}%
\hfill
\begin{subfigure}[b]{0.2099\textwidth}
    \centering
    \includegraphics[width=\linewidth]{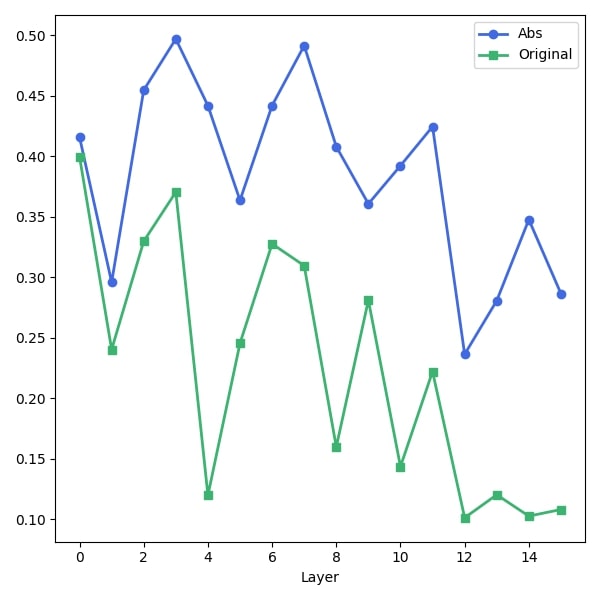}
    \vspace{-1.7em}
    \caption{Pearson's Corr.}
    \label{fig:magnitude:b}
\end{subfigure}
\vspace{-0.3em}
\caption{Correlation between Llama attention and \abbr~attention.}
\vspace{-0.8em}
\label{fig:magnitude}
\end{wrapfigure}

In Fig.\ref{fig:magnitude}, we present the correlation between attention scores from Llama and \diff, computed layer by layer. Specifically, we compare Llama's softmax scores against both the original signed scores from \abbr~attention ({\color{mediumseagreen}green}) and their absolute values ({\color{royalblue}blue}). Note that while standard Llama attention scores are non-negative (due to softmax), \abbr~attention scores can be negative. Interestingly, both rank and Pearson correlations are significantly higher when using absolute values ({\color{royalblue}blue}) compared to signed values ({\color{mediumseagreen}green}). This suggests strong correspondence in the \textit{magnitude} of attention (indicating relative importance), even when the signed scores differ. This observation motivates our \dex~design: since the relative importance signals (magnitudes) from the standard QK/softmax pathway are largely preserved, we reuse them and focus our adaptation efforts on enhancing the subsequent OV circuit to incorporate differential mechanism.

\subsection{Attention Head Redundancy}
\label{supp:2.2}

We address the potential concern that lower inter-head redundancy in \diff~stems from its common configuration using fewer, wider attention heads (typically halving head count while doubling head dimension~\footnote{\url{https://github.com/microsoft/unilm/issues/1663}}).
% might be attributed to \diff~having fewer heads (note that \diff~chooses to double the head dimension while halving the number of heads).

\begin{wrapfigure}{r}{0.45\textwidth}
\vspace{-1.6em}
\centering
\setlength\intextsep{0pt}
\resizebox{0.41\textwidth}{!}{%
    \includegraphics[width=\textwidth]{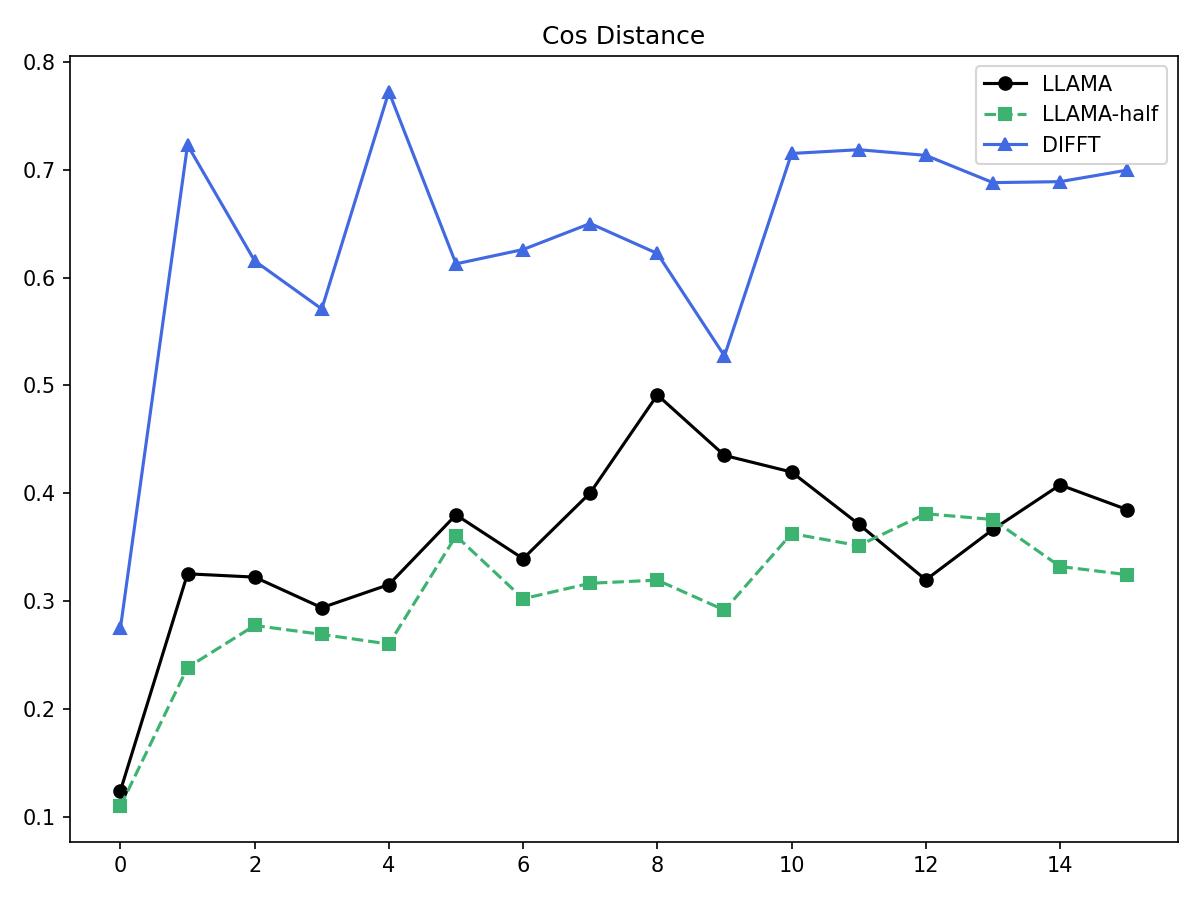}
}
\caption{Mean pairwise cosine distance between attention scores from different heads. }
\vspace{-1.4em}
\label{fig:wide}
\end{wrapfigure}

% We plot the cosine distance between attention scores from attention heads within each layer, computed pairwise and averaged (Fig.\ref{fig:wide}). 
% The figure clearly shows that \abbr~attention exhibits much higher average cosine distance across all layers, indicating that the redundancy between attention heads is lower (attention patterns are more dissimilar). 
We plot the average pairwise cosine distance between head attention scores per layer (Fig.\ref{fig:wide}).
The figure shows \abbr~attention exhibiting significantly higher average cosine distance, indicating lower redundancy (greater pattern dissimilarity) among its heads.
Notably, merely using fewer, wider heads does not replicate this effect, as demonstrated by our LLAMA-half baseline ({\color{mediumseagreen}green}), configured with halved head count and doubled head dimension. 
We hypothesize that the differential mechanism grants greater flexibility in controlling attention patterns, thus reducing inter-head redundancy.
% We hypothesize that the differential mechanism provides additional knobs to control the attention pattern more flexibly, leading to reduced overall redundancy among attention heads.

\begin{figure}[h]
    % \vspace{-1em}
    \centering
    \includegraphics[width=0.9\textwidth]{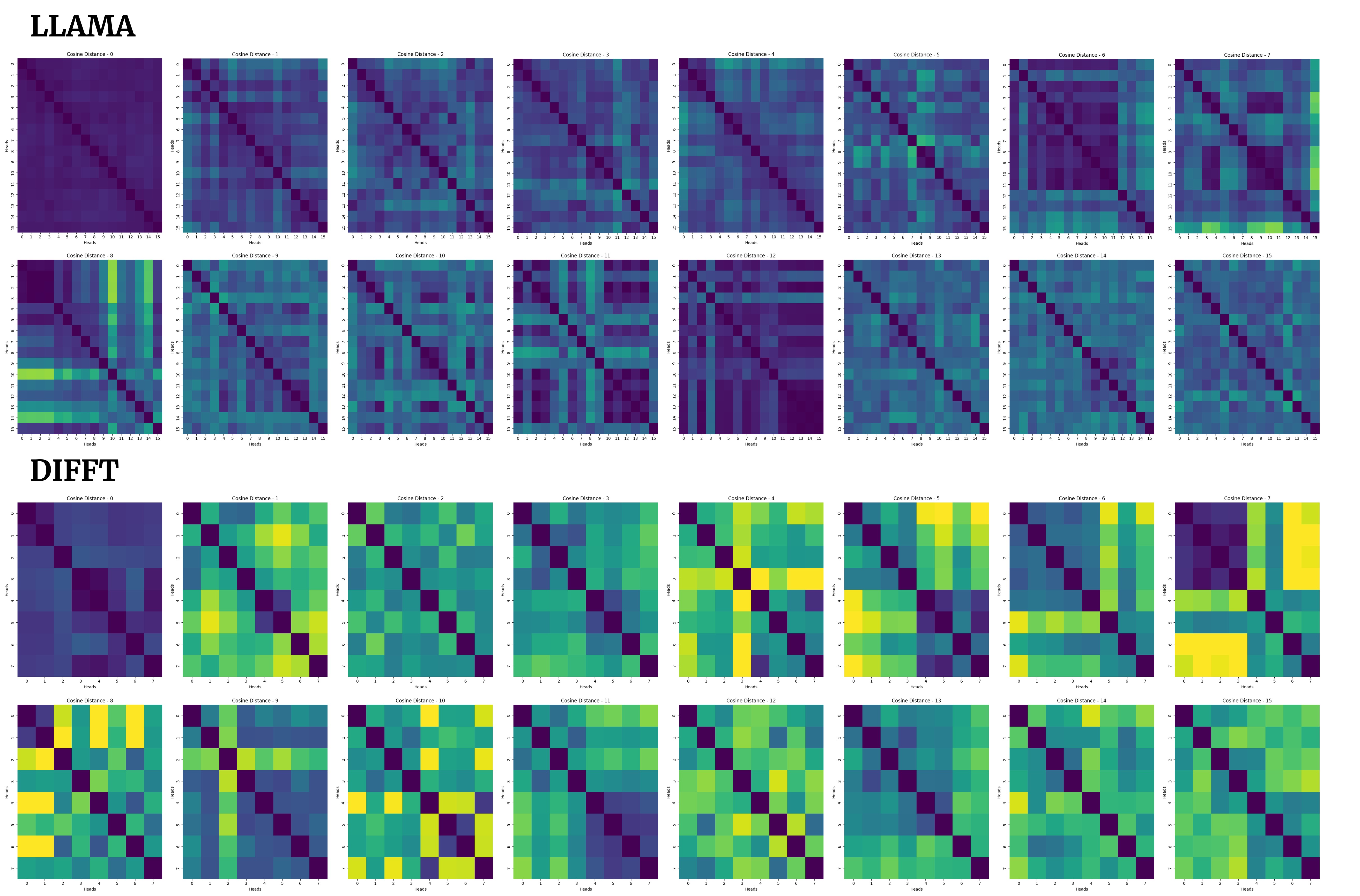}
    \captionsetup{font=footnotesize} % or scriptsize, footnotesize, etc.
    \caption{Pairwise cosine distance between attention maps from different attention heads in each layer. Brighter color indicates larger distance, hence lower redundancy.}
    \label{fig:supp_cos}
    % \vspace{-1em}
\end{figure}

Heatmaps visualizing the pairwise cosine distances between attention maps from different heads (Fig.\ref{fig:supp_cos}) further corroborate our findings. They show lower inter-head distances (indicating higher similarity and redundancy) in the standard transformer (LLAMA), whereas \diff~maintains higher distances, demonstrating more diverse attention patterns.

\begin{figure}[h]
    \centering
    \includegraphics[width=\textwidth]{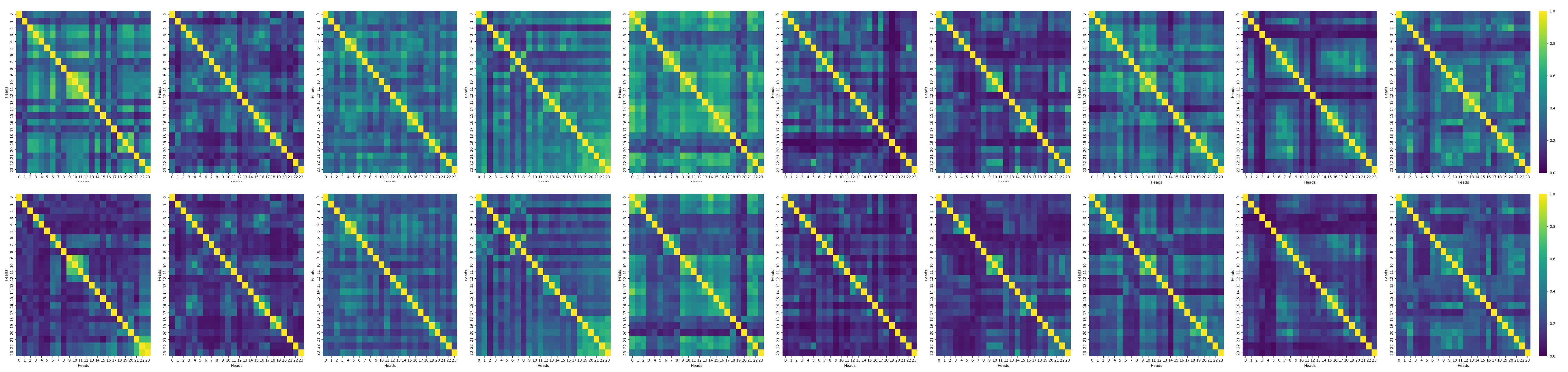}
    \captionsetup{font=footnotesize} % or scriptsize, footnotesize, etc.
    \caption{Centered Kernel Alignment for attention heads. Brighter colors indicate higher alignment/similarity. (Top) Llama, (Bottom) \dex.}
    \label{fig:supp_cka}
    \vspace{-1em}
\end{figure}

Centered Kernel Alignment (CKA) analysis comparing attention heads before and after applying \dex~(Fig.\ref{fig:supp_cka}) further confirms that \dex~reduces inter-head redundancy. The results clearly show lower overall alignment between heads after adaptation in the pretrained models, indicating increased functional diversity.

\pagebreak

\subsection{Abalation on $\mathbf{\lambda_{init}}$}
\label{supp:2.3}

\begin{wraptable}{r}{0.45\textwidth} 
    \centering 
    \vspace{-1.3em} 
    \caption{Ablation on $\lambda_{init}$. \abbr~refers to depth-aware initialization following \cite{ye2024differential}.}
    \label{tab:lambda_abl} 
    \resizebox{0.45\textwidth}{!}{
        \begin{tabular}{l | c c c c} 
            \specialrule{1.5pt}{0pt}{0.3em} 
            % \vspace{0.5em}            
            $\lambda_{init}$ & 0.8 & 0.5 & 0.3 & \abbr  \\
            \midrule 
            LM Acc (\%) & 54.3 & 54.0 & 54.2 & \textbf{54.4}  \\
            \bottomrule
        \end{tabular}}
    \vspace{-0.3em} 
\end{wraptable}

Table \ref{tab:lambda_abl} shows \dex~performance on the language modeling benchmarks (average over 11 tasks from Table~\ref{tab:main}, using Qwen-0.5B) when varying the $\lambda_{init}$ strategy. 
% While it is clear that the performance is relatively robust to the specific choice or $\lambda_{init}$ values, depth-aware initialization (adopted from \cite{ye2024differential}) yields the best result. 
The results indicate relative robustness to different fixed scalar initializations (0.3-0.8). However, adopting the initialization scheme from the original \diff~setting yields slightly the best performance.
We hypothesize the layer-aware initialization is beneficial for training.

\subsection{Ablation on Head Selection $k$}
\label{supp:2.4}

\begin{wraptable}{r}{0.45\textwidth} 
    \centering 
    \vspace{-1em} 
    \caption{Ablation on head selection $k$.}
    \label{tab:head_k} 
    \resizebox{0.45\textwidth}{!}{
        \begin{tabular}{l | c c c c} 
            \specialrule{1.5pt}{0pt}{0.3em} 
            % \vspace{0.5em}            
            $k$ & 8 & 16 & 24 & 32   \\
            \midrule 
            LM Acc (\%) & 53.7 & \textbf{55.6} & 54.4 & 54.1  \\
            \bottomrule
        \end{tabular}}
    \vspace{-1em} 
\end{wraptable}

In Table \ref{tab:head_k}, we present the average performance of \dex~with different number of target attention heads ($k$) on 11 language modeling benchmarks. 
% Selecting too little number of heads (\textit{e.g.,} $k=8$) gives insufficient flexibility for \dex~to assert its impact, while modifying excessive amount of heads leads to performance degradation as some of the critical pretrained knowledge can be affected. 
Selecting too few heads (\textit{e.g.,} $k=8$) provides insufficient capacity for the differential adaptation, while modifying too many heads risks disrupting critical pretrained knowledge, leading to performance degradation.
We empirically find modifying about 50\% of the attention heads tends to be optimal in general (note we use Llama-1B with 32 heads for ease of demonstration).

\subsection{Qualitative Results}
\label{supp:2.5}

We display additional examples from qualitative analysis in Sec.\ref{sec:2.2}.

\begin{figure}[h]
    \centering
    \includegraphics[width=\textwidth]{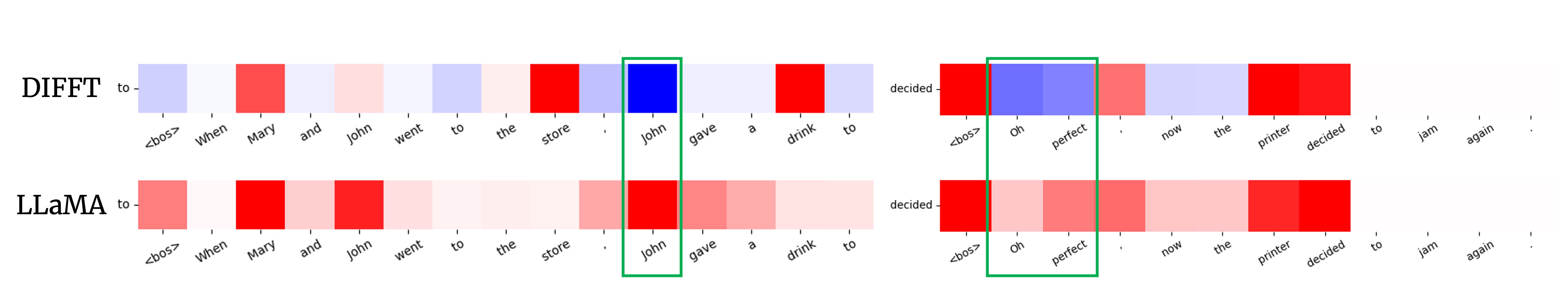}
    \captionsetup{font=footnotesize} % or scriptsize, footnotesize, etc.
    \caption{Qualitative examples for \diff~and Llama.}
    \label{fig:supp_qual}
    % \vspace{-1em}
\end{figure}

% Fig.\ref{fig:supp_qual} is generally in line with Fig.\ref{fig:qual}, demonstrating how \abbr~attention leverages negative attention to form richer representations. The example in the left shows an Indirect Object Identification task, where \diff~intentionally assigned negative attention score to mark the subject (\textit{i.e.,} John) as irrelevant. The one in the right shows another example of sarcasm, where \abbr~attention cleverly identifies non-literal expression and explicitly allocate negative attention score.
Fig.\ref{fig:supp_qual} aligns with the observations from Fig.\ref{fig:qual}, illustrating how \abbr~attention leverages negative attention scores. The left example shows an Indirect Object Identification task where \diff~assigns a negative attention score to mark the subject (\textit{i.e.,} John) as irrelevant. The right example shows sarcasm detection, where \abbr~attention identifies the non-literal expression and explicitly allocates negative attention scores accordingly.

\section{Comparison to \diff}
\label{supp:4}

A direct comparison with the original \diff~model is not possible due to unavailable weights. Therefore, to establish a point of reference, we compare \dex~with a \diff~model that we trained ourselves at a smaller scale, following the procedures detailed in Appendix \ref{supp:5.2}.
To set up this comparison, we first train two models from scratch on the exact same training data: (1) a standard transformer baseline using the Llama architecture (Llama), and (2) \diff~model (\abbr). Subsequently, we apply \dex~to the Llama baseline using a small subset of the pretraining data (<1B tokens) to create the third model, simply noted \dex. 
We additionally train a separate Llama model from scratch with \dex~attached from the beginning, to understand \dex's architectural capacity beyond its original purpose of adaptation, which we refer to as \dex-S.
We report the performance of these four models (Llama, \abbr, \dex, \dex-S) on the 11 language modeling benchmarks in Table~\ref{tab:supp_lm}.

\begin{table}[h]
\vspace{-1em}
\centering
\caption{Scores on 11 benchmarks. \colorbox{win}{Green} indicates increases and \colorbox{lose}{gray} indicates decreases. All values are rounded to one decimal place. 
}
\label{tab:supp_lm}
\resizebox{\textwidth}{!}{%
\begin{tabular}{lccccccccccc|cc}
\toprule
\textbf{Model} & \textbf{Arc-C} & \textbf{Arc-E} & \textbf{BoolQ} & \textbf{COPA} & \textbf{Hellaswag} & \textbf{MNLI} & \textbf{OBQA} & \textbf{PIQA} & \textbf{WIC} & \textbf{Winogrande} & \textbf{WSC} & \textbf{AVG} & \textbf{$\Delta$} \\
\regmidrule
\\[-0.8em]
Llama       & 21.8 & 37.0 & 60.5 & 63.0 & 29.0 & 35.1 & 25.2 & 58.4 & 50.6 & 49.6 & 36.5 & 42.4 & - \\
\\[-0.8em]
\abbr     & \cellcolor{win}24.2 & \cellcolor{win}37.2 & \cellcolor{lose}54.0 & \cellcolor{win}68.0 & \cellcolor{win}29.0 & \cellcolor{win}35.5 & \cellcolor{win}26.4 & \cellcolor{win}58.9 & \cellcolor{lose}50.0 & \cellcolor{win}52.2 & \cellcolor{win}36.5 & \cellcolor{win}42.9 & \textcolor{deltaGreen}{+0.5} \\
\\[-0.8em]
\dex   & \cellcolor{win}22.2  & \cellcolor{win}37.1  & \cellcolor{win}60.5  & \cellcolor{win}64.0  & \cellcolor{win}29.0  & \cellcolor{win}35.1  & \cellcolor{win}25.8  & \cellcolor{lose}58.3  & \cellcolor{win}50.6  & \cellcolor{win}51.2  & \cellcolor{win}36.5  & \cellcolor{win}42.8  & \textcolor{deltaGreen}{+0.4} \\
\\[-0.8em]
\dex-S   & \cellcolor{win}22.5  & \cellcolor{win}37.1  & \cellcolor{win}61.5  & \cellcolor{win}63.0  & \cellcolor{lose}28.7  & \cellcolor{win}35.2  & \cellcolor{win}27.4  & \cellcolor{lose}58.1  & \cellcolor{lose}50.0  & \cellcolor{win}51.0  & \cellcolor{win}36.5  & \cellcolor{win}42.8  & \textcolor{deltaGreen}{+0.4} \\
\regmidrule
\end{tabular}
}
\vspace{-0.8em}
\end{table}

From the table, we first observe that \diff~generally outperforms standard transformer on the majority of benchmarks, which supports the strength of \diff~as a general purpose language model. Furthermore, the results clearly demonstrate that \dex, despite being lightweight both during training and inference, effectively enhances the pretrained Llama model, closing the gap between standard transformer and \diff. \dex-S, a variant of \dex~applied from scratch, also delivers competitive performances beyond standard Llama model.

\begin{figure}[h]
    \centering
    \includegraphics[width=\textwidth]{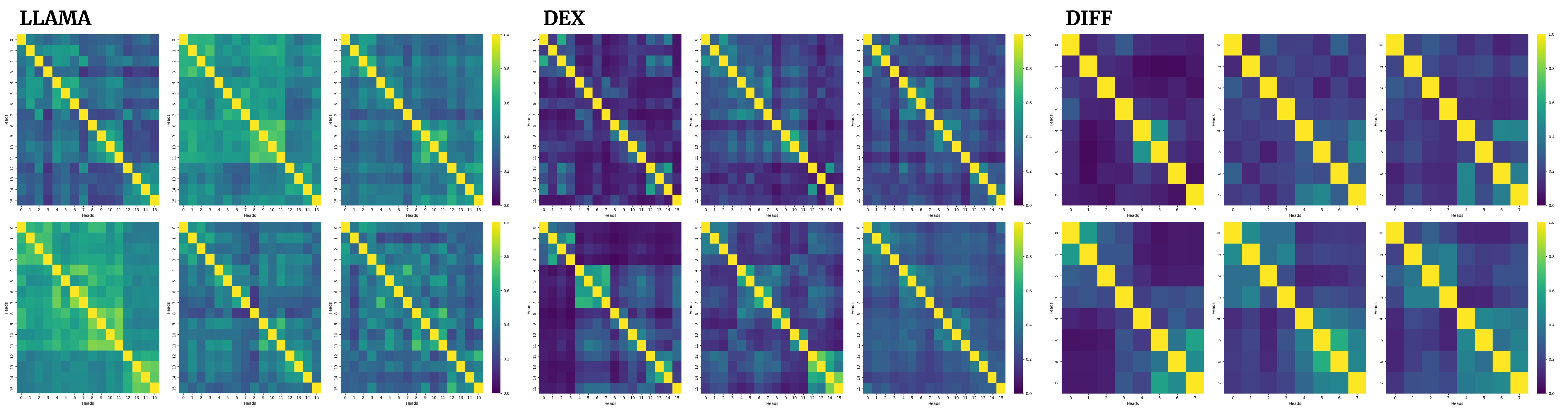}
    \captionsetup{font=footnotesize} % or scriptsize, footnotesize, etc.
    \caption{Head CKA comparison between Llama vs \dex~vs \abbr.}
    \label{fig:supp_head_cka}
    % \vspace{-1em}
\end{figure}

% TODO: start from here
% HEAD-CKA figure goes here

Head CKA results further support the effectiveness of \dex~(Fig.\ref{fig:supp_head_cka}). Compared to standard transformer (left), \dex~significantly reduces the inter-head redundancy (indicated by lower alignment), yielding similar results to \diff.

\section{Efficiency Analysis}
\label{supp:efficiency}

To evaluate inference efficiency, we benchmark throughput (tokens per second) for 3B-parameter versions of Llama, \diff, and \dex, presenting the results in Fig.\ref{fig:throughput}. Context lengths were varied from 1k to 64k tokens to cover a comprehensive range of use cases. All tests were conducted on a single NVIDIA A100-80GB GPU, utilizing PyTorch's standard scaled dot-product attention implementation\footnote{\url{https://docs.pytorch.org/docs/stable/generated/torch.nn.functional.scaled_dot_product_attention.html}}. The reported throughputs are averaged over 30 batches, following an initial 5 warm-up batches.

\section{Implementation Details}
\label{supp:5}

In this section, we provide comprehensive details for our experiments, some of which were abbreviated in the main manuscript for brevity.

\subsection{Language Modeling Evaluation}
\label{supp:5.1}

\paragraph{Dataset}
We constructed our custom training corpus using a subset of the Dolmino dataset\footnote{\url{https://huggingface.co/datasets/allenai/dolmino-mix-1124}}. 
Specifically, we mixed web pages, academic papers, encyclopedia entries, and code texts in approximate ratios of 74.3\%, 6.5\%, 7.9\%, and 11.3\% respectively. This resulted in a corpus totaling 887M tokens (measured using the Llama-3 tokenizer).
Our data preparation generally followed the recipe of OLMo2~\cite{olmo20242}, with the main exception being a greater upsampling of the code text component.

\paragraph{Training}
% We train both the baselines and \dex~with our custom corpus for 1 epoch. 
All models, including baselines and \dex~variants, were trained on our custom corpus for 1 epoch.
% We set context length to be $32k$ for both Llama and Qwen. 
A context length of $32k$ tokens was used for all Llama and Qwen models during this training phase.
% We use cosine schedule with learning rate 1e-4 for partial fine-tuning methods (including \dex) and 1e-5 for full fine-tuning baselines, as this generally resulted in best outcome. We employ warm up ratio of 0.03.
We employed a cosine learning rate schedule, using a peak learning rate of $1 \times 10^{-4}$ for partial fine-tuning methods (including \dex) and $1 \times 10^{-5}$ for full fine-tuning baselines, as these settings generally yielded the best outcomes in preliminary experiments.
A learning rate warm-up ratio of 0.03 was used. 
% All experiments were conducted using 8 NVIDIA A100-80GB GPUs, with the run time ranging from 2.5-16 hours depending on the model size.

\subsection{Training \diff}
\label{supp:5.2}

We train our own \diff~model for analysis. This subsection details its training procedure.

\paragraph{Dataset}

We followed the recipe of \diff~and StableLM-3B\footnote{\url{https://github.com/Stability-AI/StableLM}}, using various open-source datasets~\cite{refinedweb,gao2020pile,li2023starcoder,weber2024redpajama} to create a corpus of approximately 30 billion tokens (Llama-3 tokenizer). This corpus encompasses a diverse range of domains, including academic papers, source code, encyclopedic articles, and literature.

\paragraph{Model}

We trained a 0.4-billion parameter version of \diff. Key architectural parameters are provided in Table~\ref{tab:config}.

\begin{table}[h]
    \centering 
    \caption{Configuration for 0.4B \diff.}
    \label{tab:config} 
    \resizebox{0.25\textwidth}{!}{%
        \begin{tabular}{l | c } 
            \toprule 
            params & values \\
            \midrule 
            \# Layers & 16 \\ 
            \# Heads & 16 \\ 
            \# KV Heads & 4 \\ 
            Hidden size & 1024 \\
            FFN size & 4096 \\
            \bottomrule
        \end{tabular}%
    }
\end{table}

\paragraph{Training}

For training, we employed the AdamW optimizer~\cite{loshchilov2017decoupled} with a cosine learning rate schedule.
The peak learning rate was set to $1 \times 10^{-4}$, the global batch size to 256, and the learning rate warm-up to 0.1.
% We initialized lambda using the exact schedule of \diff.
The $\lambda$ parameters within the differential attention were initialized according to the exact schedule specified in the original \diff~paper~\cite{ye2024differential}.

\subsection{Approximating Effective Attention Scores for \dex~Interpretability}
\label{supp:5.3}

Because \dex~directly alters the attention block's output value matrix $\mathbf{O}$ rather than the initial softmax scores, standard attention visualization can be misleading. To provide insight into its effective learned behavior, we propose methods to approximate the \textit{effective attention scores} that would yield \dex's modified output using the original value matrix.

\paragraph{Least-Squares Approximation}
This method uses the Moore-Penrose pseudoinverse to derive effective attention scores $\mathbf{X}$ that best reconstruct \dex's output transformation.
% Specifically, let $A = \text{softmax}(QK^T)$ be the attention scores (we omit the scaling factor for brevity), $V \in \mathbb{R}^{n\times d}$ be the value matrix, and $D \in \mathbb{R}^{d\times d}$ be the weight matrix for $f_D$ in \dex. 
Specifically, let $\mathbf{A} = \text{softmax}(\mathbf{QK}^T/\sqrt{d_k})$ be the original softmax attention scores from a given head, $\mathbf{V} \in \mathbb{R}^{N \times d_v}$ be the corresponding value matrix (where $N$ is sequence length, $d_k$ is key dimension, $d_v$ is value dimension), and $\mathbf{W}_D \in \mathbb{R}^{d_v \times d_v}$ be the learnable weight matrix for $f_D$ in \dex~(assuming $\lambda(t)$ is absorbed into $\mathbf{W}_D$ or considered $\approx 1$ for this analysis).
% We want to compute the \textit{effective value weights} $X$ such that
The original head output is $\mathbf{O} = \mathbf{A}\mathbf{V}$, and the \dex-modified output is $\mathbf{O}' = \mathbf{O}(I - \mathbf{W}_D)$. We seek an effective attention score matrix $\mathbf{X}$ such that $\mathbf{X}\mathbf{V} \approx \mathbf{O}'$.

% \begin{equation}
% \label{eq:pinv}
%     AV - AVD = XV.
% \end{equation}

The least-squares solution for $\mathbf{X}$ is:
\begin{equation}
    \mathbf{X} = \mathbf{O}' \mathbf{V}^+ = \mathbf{A}\mathbf{V}(I-\mathbf{W}_D)\mathbf{V}^+
    \label{eq:eff_attn_scores} % More descriptive label
\end{equation}
where $\mathbf{V}^+$ denotes the Moore-Penrose pseudoinverse of $\mathbf{V}$, computed numerically in practice.

This $\mathbf{X}$ represents the attention pattern that, if applied to the original values $\mathbf{V}$, would best reconstruct \dex's modified output for that head. Since this involves an approximation and the use of a pseudoinverse (which can be sensitive if $\mathbf{V}$ is ill-conditioned or has a significant null space), numerical considerations are important. We therefore complement and cross-check these results using a second technique.

\paragraph{Optimization-based Approximation}
As with the pseudoinverse method, we aim to find an effective attention score matrix $\mathbf{X}$ such that $\mathbf{X}\mathbf{V}$ approximates the \dex~output $\mathbf{O}' = \mathbf{A}\mathbf{V}(I-\mathbf{W}_D)$. Rather than a closed-form pseudoinverse solution, this approach directly optimizes for $\mathbf{X}$ for each input sample for which $\mathbf{O}'$ and $\mathbf{V}$ are computed. For each sample, $\mathbf{X}$ is typically initialized (e.g., as the original attention scores $\mathbf{A}$) and then updated for 100 iterations using gradient descent (learning rate $1 \times 10^{-3}$) to minimize a reconstruction loss with the form $||\mathbf{X}\mathbf{V} - \mathbf{O}'||_2^2$.

The primary interpretable attention scores reported in our main analyses (e.g., Table~\ref{tab:noise_cancel}) were derived using the pseudoinverse method. This optimization-based approach served as a cross-validation, and we confirmed strong agreement between the effective attention scores obtained from both techniques. While both methods yield approximations subject to numerical precision, they offer valuable tools for understanding the internal mechanisms and effective attention patterns of \dex.

\section{Broader Impact}
\label{supp:broad}

\textbf{Potential Positive Societal Impacts:} By improving core LLM capabilities such as information retrieval, in-context learning, and overall representational quality, \dex~could contribute to more effective and reliable AI systems. This includes advancements in AI-assisted education, more capable research tools, improved accessibility to information, and more helpful AI assistants. Furthermore, DEX's design emphasizes lightweight adaptation, which could make powerful LLM enhancements more resource-efficient and accessible, potentially reducing the computational burden associated with adapting large models.

\textbf{Potential Negative Societal Impacts:} As \dex~is designed to improve the capabilities of LLMs, it shares the potential negative societal impacts inherent in more powerful language model technology. Enhancements in LLM performance and efficiency could inadvertently facilitate the creation of more sophisticated or scalable misuse scenarios, such as generating convincing disinformation, spam, or impersonations. If an LLM enhanced by \dex~produces incorrect or biased information, its improved fluency might make such outputs seem more authoritative, potentially exacerbating harm. While \dex~is a foundational architectural improvement rather than a specific end-user application, the dual-use nature of advancements in LLM capabilities warrants careful consideration.

We believe that continued research into robust AI safety measures, ethical development guidelines, bias detection and mitigation, and responsible deployment practices for all LLMs is crucial as their capabilities, including those enhanced by methods like \dex, advance.

\end{document}